\documentclass[journal]{IEEEtran}
\usepackage[linesnumbered,ruled,vlined]{algorithm2e}
\usepackage{multirow,booktabs,color,soul,threeparttable,longtable}
\usepackage{booktabs}
\usepackage{multirow}
\usepackage{hyperref}
\hypersetup{hidelinks} 
\usepackage{amsmath,amsfonts}
\usepackage{subfigure}
\usepackage{epsfig}
\usepackage{array}
\usepackage{textcomp}
\usepackage{stfloats}
\usepackage{url}
\usepackage{verbatim}
\usepackage{graphicx}
\usepackage{cite}
\usepackage{amssymb}
\usepackage {color}
\ifCLASSINFOpdf
\else
\fi
\begin{document}
	%
	\title{REMS: a unified solution representation, problem modeling and metaheuristic algorithm design for general combinatorial optimization problems}
	%
	%
	%
	
	\author{Aijuan~Song,
		Guohua~Wu~\IEEEmembership{Senior Member,~IEEE}
		
		 \thanks{Accepted for publication at IEEE/CAA Journal of Automatica Sinica. The final version may differ.}
		 
		\thanks{This work was supported by the National Natural Science Foundation of China 62373380, the Fundamental Research Funds for the Central Universities of Central South University under Grant 2025ZZTS0108, the Natural Science Foundation of Hunan Province under Grant 2025JJ10007, and the National Natural Science Foundation of China under Grant 62403493. (\emph{Corresponding author: Guohua Wu.})}
		
		\thanks{Aijuan Song is with the School of Traffic and Transportation Engineering, Central South University, Changsha 410075, China. E-mail: aijuansong@csu.edu.cn.}
			
		 \thanks{Guohua Wu is with the School of Automation, Central South University, Changsha 410083, China. E-mail: guohuawu@csu.edu.cn.}
		
	}
	
	%
	%

	\markboth{Accepted for publication at IEEE/CAA Journal of Automatica Sinica. The final version may differ.}%
	{XX \MakeLowercase{\textit{et al.}}: Bare Demo of IEEEtran.cls for IEEE Journals}
	%



	\maketitle
	
	\begin{abstract}
		Combinatorial optimization problems (COPs) with discrete variables and finite search spaces are critical across various fields, and solving them in metaheuristic algorithms is popular. However, addressing a specific COP typically requires developing a tailored and handcrafted algorithm. Even minor adjustments, such as constraint changes, may necessitate algorithm redevelopment. Therefore, \textcolor{black}{it is valuable to leverage general problem domain knowledge to establish} a framework that formulates diverse COPs into a unified paradigm and supports the design of broadly applicable metaheuristic algorithms. A COP \textcolor{black}{can typically be} viewed as the process of giving resources to perform specific tasks, \textcolor{black}{subject} to given constraints. Motivated by this, a resource-centered modeling and solving framework (REMS) is introduced. We first extract and define resources and tasks from a COP. Subsequently, given predetermined resources, the solution structure is unified as assigning tasks to resources, from which variables, objectives, and constraints can be derived, thereby constructing the problem model. To solve the COPs, several fundamental operators are designed \textcolor{black}{from the resource-task perspective} based on the unified solution structure, including the initial solution, neighborhood structure, destruction and repair, crossover, and ranking. These operators enable the development of various metaheuristic algorithms. \textcolor{black}{Specifically}, 4 single-point-based algorithms and 1 population-based algorithm are configured herein. Experiments on 10 COPs, covering routing, location, loading, assignment, scheduling, and graph coloring problems, show that REMS can model these COPs within the unified paradigm and effectively solve them by the algorithms in REMS \textcolor{black}{without any specific design}. Furthermore, REMS is more competitive than GUROBI and SCIP in tackling large-scale instances and complex COPs, and outperforms OR-TOOLS on several challenging COPs.
	\end{abstract}
	

	%
	\IEEEpeerreviewmaketitle

	\section{Introduction}
	%
	%
	%
	%
	\IEEEPARstart{C}{ombinatorial} optimization problems (COPs) with discrete decision variables and finite search spaces are widely encountered and hold significant importance across various fields \cite{du1998handbook}, including logistics, engineering, \textcolor{black}{economics}, and healthcare. A COP can be formulated as \cite{karimi2022machine}
	\begin{equation}\label{Eq:COP}
	\begin{array}{l}
	{\rm{min     }} {\kern 5pt} f(\vec x)\\
	{\rm{s.t.   }}{\kern 5pt}G(\vec x) \le 0,{\kern 4pt}H(\vec x) = 0
	\end{array},
	\end{equation}
	where $\vec x = \{ {x_1}, \cdots ,{x_n}\}$ denotes variables, ${x_i} \in \Omega_i \subset {\mathbb{Z}^n},{\kern 4pt}i \in \{ 1, \cdots ,n\} $, and $\Omega_i$ is a discrete and finite variable space for $x_i$. $G(\vec x) \le 0$ and $H(\vec x) = 0$ are the inequality and equality constraints, respectively. Put all possible feasible solutions into $\mathcal{F}$, then solving such a COP is to find a solution $\vec x^{*} \in \mathcal{F}$ to enable $f(\vec x^{*}) \le f(\vec x),{\kern 4pt}\forall \vec x \in \mathcal{F}$.
	
	In the real world, addressing evolving needs and requirements necessitates adapting variables $\vec x$, objective functions $f(\vec x)$, as well as constraints $G(\vec x)$ and $H(\vec x)$. It gives rise to various COPs such as routing \cite{li2022overview}, location \cite{lei2023chaotic}, loading \cite{ekici2023large}, assignment \cite{krumke2013generalized}, scheduling \cite{dauzere2024flexible}, and graph coloring problems \cite{jensen2011graph}, along with their variants \cite{kaya2022review}. Most of these problems are considered NP-hard \cite{kaya2022review, karimi2022machine}.
	
	Researchers have dedicated considerable effort to \textcolor{black}{constructing} an appropriate model for a COP under a specific context and to \textcolor{black}{designing} tailored algorithms for solving it, such as exact and heuristic methods \cite{marti2022exact}. In addition, to overcome the limitation of problem-specific formulations and algorithm-tailored methods, expressing a broad range of COPs as a unified paradigm and developing more general algorithms to address them are important pursuits \cite{pagnozzi2019automatic, kallestad2023general, marti202450}, which \textcolor{black}{facilitate} the emergence of various generic solvers and related research.
	
	Exact algorithm-based solvers, such as GUROBI, CPLEX, SCIP, COIN-OR, and CBC, already possess the capability to express various COPs as unified paradigms and solve them without requiring the redesign of algorithms. They generally formulate a COP into a strictly defined model structure of mixed integer programming (MIP). Subsequently, an exact algorithm, like a branch-and-bound method \cite{lawler1966branch}, is mainly employed to solve the constructed model. The solvers typically possess good performance in simple and small-scale COPs. However, certain challenges persist. First, some problems, such as vehicle routing problems, may lack a well-structured mathematical formulation \cite{marti202450}. Modeling for such COPs requires advanced expertise. Second, these solvers have limited ability to model and address COPs with complex nonlinear operators, such as logarithmic, exponential, and power functions, in the objective functions or constraints. Furthermore, the performance of exact algorithms themselves deteriorates as problem complexity and scale increase. 
	
	Heuristic algorithms employ dedicated rules developed by extracting problem domain knowledge to determine reasonable solutions within an acceptable computational burden. They are generally considered more suitable for large-scale and complex problems than exact methods, and can be broadly categorized into classical heuristic and metaheuristic algorithms. Classical heuristic algorithms include Greedy \textcolor{black}{Algorithm} \cite{zhao2021iterated}, Clarke-Wright Algorithm  \cite{nelson1985implementation}, and Sweep Algorithm \cite{laporte2000classical}, while metaheuristic algorithms encompass Simulated Annealing (SA) \cite{tian1999application}, Tabu Search (TS) \cite{chelouah2000tabu}, Large Neighborhood Search (LNS) \cite{mara2022survey}, Variable Neighborhood Search (VNS) \cite{brimberg2023variable}, and Genetic Algorithm (GA) \cite{zhao2024pega}. Metaheuristic algorithms generally possess better global search capabilities than classical heuristic methods \cite{peres2021combinatorial}.
	
	When addressing a COP by heuristic algorithms, an overwhelming majority of existing research still concentrates on developing dedicated algorithms tailored to specific COPs. The process is typically tedious and time-consuming. \textcolor{black}{Moreover, due to the limitation of specific problem domain knowledge,} even if a minor variation to a COP, such as changes in constraints, can make the original algorithm inefficient or even inappropriate, leading to a redesign of the algorithm.
	
	\textcolor{black}{Given the advantages and challenges in heuristic methods, several solvers have been developed by integrating them to address various problems.} Google's CP-SAT in OR-TOOLS \cite{moreira2024cp}, Hexaly, and OptaPlanner are notable examples. Among them, CP-SAT formulates problems as constraint programming models using a dedicated modeling language \cite{thakurani2024leveraging}. \textcolor{black}{It solves these problems based on the Lazy Clause Generation (LCG) method \cite{bit2023enhancing}, while integrating various heuristic strategies, such as the Feasibility Pump and LNS, to explore the solution space \cite{cuvelier2023or}. Hexaly introduces list variables to improve modeling. It is suitable to solve \textcolor{black}{problems} with sequence decision structures, including routing and scheduling problems \cite{marti202450}. Hexaly also supports various nonlinear operators and optimizes problems by exact and heuristic methods \cite{blaisemodeling}. OptaPlanner employs object-oriented modeling by defining problem facts, planning entities, variables, and both soft and hard constraints \cite{zhang2023balancing}. It applies a penalty function to assess solutions and provides various heuristic methods, including TS, LNS, and VNS, to be selected based on problem characteristics.} Notably, some exact algorithm-based solvers, like GUROBI and CPLEX, also integrate heuristic rules to enhance performance.
	
	Additionally, there are several solvers \textcolor{black}{tailored} for a single type of COP. For instance, the Routing Solver \cite{cuvelier2023or} in OR-TOOLS specializes in Vehicle Routing Problems (VRPs), providing modeling and heuristic rules based on the specific problem knowledge. Furthermore, certain researchers have devoted themselves to designing general and effective heuristic algorithms for one particular type of COPs, such as the research in \cite{pagnozzi2019automatic, vidal2014unified, blum2016construct, sarafraz2021unified, shi2024fixed}. 
	
	Despite these efforts, research on problem representation and general-purpose heuristic algorithms remains limited, as seen in \cite{marti202450}, and still encounters challenges. For example, although certain solvers, such as CP-SAT and OptaPlanner, can handle some nonlinear operators, they struggle to effectively model and solve COPs with highly nonlinear operators, such as logarithmic, exponential, and power functions. \textcolor{black}{In addition, solvers like CP-SAT and OptaPlanner exhibit limited capability when modeling and dealing with problems containing continuous variables.} 
	
	\textcolor{black}{It is also worth noting that the application of reinforcement learning (RL) and neural networks to solve COPs, known as Neural Combinatorial Optimization (NCO) \cite{kwon2020pomo, wu2024neural, bello2016neural}, has gained considerable attention. The NCO method can derive an effective agent through extensive training, which, once learned, can be applied to solve problems rapidly. Moreover, it can be generalized to the COPs with similar structural characteristics and data distributions. However, changes in variables, objectives, or constraints may render the learned agent ineffective or inapplicable, requiring redesign or retraining.}

	\textcolor{black}{We find that most COPs can naturally be viewed as the process that gives resources to perform certain tasks. Specifically, a task refers to any assignable object (such as arcs or points in a VRP and items in a bin packing problem), while a resource is the object to which tasks are assigned (such as vehicles in the VRP and bins in the bin packing problem). With this point of view, COPs can be uniformly described as how to appropriately arrange the tasks for the resources subject to necessary constraints. Inspired by this, we propose a resource-centered modeling and solving framework (REMS) from a resource-task perspective. Compared with most tailored methods, the framework leverages more general problem domain knowledge to formulate various COPs in a unified paradigm and design broadly applicable metaheuristic algorithms to solve them.} 
	
	\textcolor{black}{Moreover, Table~\ref{tab:comparision} summarizes the differences between REMS and several widely used solvers in the aspects of modeling and solving. Concerning the modeling, REMS employs a resource-task-oriented modeling method. It is suitable for various COPs whose decisions can be reduced to assigning tasks to resources, regardless of the specific types of objectives and constraints. Regarding the algorithms, REMS focuses on applying metaheuristic methods to address COPs. Unlike black-box solvers or those limited to selecting from predefined algorithms, REMS provides better flexibility to understand, select, and customize diverse optimization strategies due to a widely adopted design process and framework of metaheuristics. In addition, compared to NCO methods, REMS can model and solve general COPs, even with changes in variables, objectives, and constraints.}
	
	\newcommand{\tabincell}[2]{\begin{tabular}{@{}#1@{}}#2\end{tabular}}
	\aboverulesep=0pt \belowrulesep=0pt
	\begin{table*}[htbp]
		\centering
		\caption{\textcolor{black}{Comparison of REMS with other solving frameworks adopted by several widely used solvers.}}
		\setlength{\tabcolsep}{0.0mm}{
			\color{black}{
				\begin{tabular}{cccccc}
				\toprule
				Solver & Modeling method & Applicable Problem Types & Modeling Flexibility & Solving Method & Algorithm Extensibility \\
				\midrule
				\tabincell{c}{CP-SAT in \\OR-TOOLS} & \tabincell{c}{Constraint \\programming} & \tabincell{c}{Problems with \\integer variables} & \tabincell{c}{Limited by highly nonlinear \\operators and continuous variables} & \tabincell{c}{Methods based on LCG\\ and heuristic rules} & \multirow{2}[1]{*}{\tabincell{c}{Low: Black-box \\optimization}} \\
				Hexaly & \tabincell{c}{Modeling with \\list variables} & \tabincell{c}{Various problems especially\\those with sequence variables} & \tabincell{c}{Support for various nonlinear \\operators and variables}  & \tabincell{c}{Exact and heuristic \\methods} & \multicolumn{1}{c}{} \\
				OptaPlanner & \tabincell{c}{Object-oriented \\modeling} & \tabincell{c}{Discrete COPs} & \tabincell{c}{Limited by highly nonlinear \\operators and continuous variables}  & Heuristic methods & \tabincell{c}{Medium: Support selecting\\ various methods} \\
				GUROBI & \multirow{3}[1]{*}{\tabincell{c}{MIP}} & \multirow{3}[1]{*}{\tabincell{c}{Problems that can be \\modeled as MIP}} & \multirow{3}[1]{*}{\tabincell{c}{Limited by highly nonlinear \\operators}}  & \multirow{3}[1]{*}{\tabincell{c}{Exact methods with \\heuristic rules}} & \multirow{3}[1]{*}{\tabincell{c}{Low: Black-box \\optimization}} \\
				CPLEX & \multicolumn{1}{c}{} & \multicolumn{1}{c}{} & \multicolumn{1}{c}{} & \multicolumn{1}{c}{} & \multicolumn{1}{c}{} \\
				SCIP  & \multicolumn{1}{c}{} & \multicolumn{1}{c}{} & \multicolumn{1}{c}{} & \multicolumn{1}{c}{} & \multicolumn{1}{c}{} \\
				\midrule
				\textbf{REMS} & \textbf{\tabincell{c}{Resource-task- \\oriented modeling}} & \textbf{\tabincell{c}{General COPs}} & \textbf{\tabincell{c}{No specific requirement on \\objectives and constraints}}  & \textbf{Metaheuristics} & \textbf{\tabincell{c}{Support selecting and \\customizing methods}} \\
				\bottomrule
			\end{tabular}%
		}
		}
		\label{tab:comparision}%
	\end{table*}%

	The main contribution of the paper is presented as follows.
	
	\begin{itemize}
		\item A resource-centered problem representation paradigm for general COPs is introduced for the first time. \textcolor{black}{By extracting resources and tasks involved in COPs,} we unify the solution structure by assigning tasks to resources within a predetermined resource set. According to the unified structure, variables, objectives, and constraints can be defined. Specifically, the variables serve to capture the specific attributes of the solution structure. The objectives are defined based on the resource-task assignment, while the constraints reflect the limitations and relationships among resources and tasks. The method \textcolor{black}{enables} representing general COPs in a unified form. 
		
		\item Several operators and metaheuristic algorithms are designed \textcolor{black}{from the resource-task perspective} to solve COPs modeled in REMS. Based on the unified solution structure, several fundamental operators are designed. An initial solution is constructed by incrementally inserting feasible tasks into resources. The operators, including neighborhood structures, destruction, repair, and crossover, are designed to search the solution space \textcolor{black}{by adjusting task assignments among resources}. Meanwhile, a hierarchical ranking method is developed \textcolor{black}{that accounts for the different scales of constraints and objectives.} These operators are integrated into single-point-based and population-based algorithm frameworks, resulting in algorithms such as SA, LNS, VNS, TS, and GA. 
		
		\item Experiments conducted on 10 COPs show that REMS can model and solve various COPs, and \textcolor{black}{exhibit} competitive performance in handling large-scale instances and complex problems than GUROBI, SCIP, and OR-TOOLS. 10 COPs from routing, location, assignment, packing, scheduling, and graph coloring problems are employed for the experimental study. The extensive experimental results reveal that REMS can effectively model and solve these problems without any specific design for algorithms. REMS outperforms GUROBI and SCIP when solving some large-scale instances and complex problems, and surpasses OR-TOOLS in addressing several problems with complex objectives and constraints.

	\end{itemize}

	The paper is organized as follows. Section~\ref{s2} details the problem representation paradigm, including motivation, framework, and key model elements. Section~\ref{s3} describes the operators and metaheuristic methods designed in REMS. Section~\ref{s4} presents the experiment to show the effectiveness of REMS. Section~\ref{s5} concludes the paper and outlines future directions.

	\section{Resource-centered problem representation method for combinatorial optimization}\label{s2}
	
	\subsection{Motivation}\label{s21}
	
	\textcolor{black}{COPs can be grouped into various categories, with each featuring different variables, objectives, and constraints. Despite their diversity, some common characteristics can be observed and extracted. In particular, we find that a COP can typically be viewed as the process of giving resources to perform specific tasks. Therefore, the core decision can be generally regarded as selecting discrete tasks from one finite task set and assigning them to discrete resources from another finite resource set. For instance, the routing problem needs to assign arcs or points to available vehicles. In the bin packing problem, items are assigned to bins. Job shop scheduling involves assigning jobs to machines, while school timetable scheduling may need to assign courses to teachers. The graph coloring problem aims at assigning colors to nodes. Accordingly, we can extract two key concepts from COPs.}
	
	\begin{itemize}
		\color{black}{\item \emph{Task}: Any assignable object, such as arc or point, item, job, course, and color, can be called a task.}
		
		\color{black}{\item \emph{Resource}: A resource is the object to which tasks are assigned. The vehicles, bins, machines, teachers, and nodes are regarded as resources in the above examples. }
		
	\end{itemize}
	
	Note that there is no clear boundary between resources and tasks for a specific COP; it depends on which type of objects need to be assigned and which type of objects they should be assigned to. For example, in a graph coloring problem, if we plan to assign colors to nodes, then the colors are assignable objects and are deemed tasks. The nodes are objects that the tasks should be assigned to and are regarded as resources. Conversely, if we plan to assign nodes to colors, then nodes can be considered tasks, while colors serve as resources.
	
	
	Furthermore, when assigning tasks to resources, the tasks should be executed in a specific order in some cases, while the execution order \textcolor{black}{is irrelevant in others}. Therefore, we assume that each resource has one or more positions to accommodate tasks, with one position dedicated to one task. These positions can either be ordered or unordered. For example, in COPs such as routing, job shop scheduling, and \textcolor{black}{school timetable scheduling}, tasks must be assigned to resources in a specific sequence to ensure sequential execution. In these cases, the resource positions are ordered and carry particular meanings, such as time slots. In contrast, for problems like bin packing and graph coloring, the order of tasks on each resource is unnecessary. The tasks simply need to be assigned to resources, and the positions are unordered.
	
	Referring to the analysis above, solving such a COP is reduced to \textcolor{black}{making} the best decision about which task to assign to which resource in which positions in order to achieve a specific objective within given constraints. If the resources are predetermined and fixed, the solution structure, which is the encoding representation that specifies how variables are organized to form a solution, can be commonly unified as determining the assigned task to a resource position. \textcolor{black}{Based on the unified solution structure, variables, objectives, and constraints can be formulated accordingly to represent various COPs. The variables capture the specific attributes of the solution structure. Moreover, the objectives are derived from the resource-task assignment, and the constraints reflect the limitations and relationships among resources and tasks. }

	\subsection{Modeling Framework and Modeling Elements}\label{s22}
	
	COPs considered in this paper have the following characteristics: (a) resources and tasks are discrete and finite; and (b) resources and tasks are predefined and remain unchanged.
	
		
		
		
	
\begin{figure}[htb]
		\begin{center}		
			\subfigure{\psfig{file=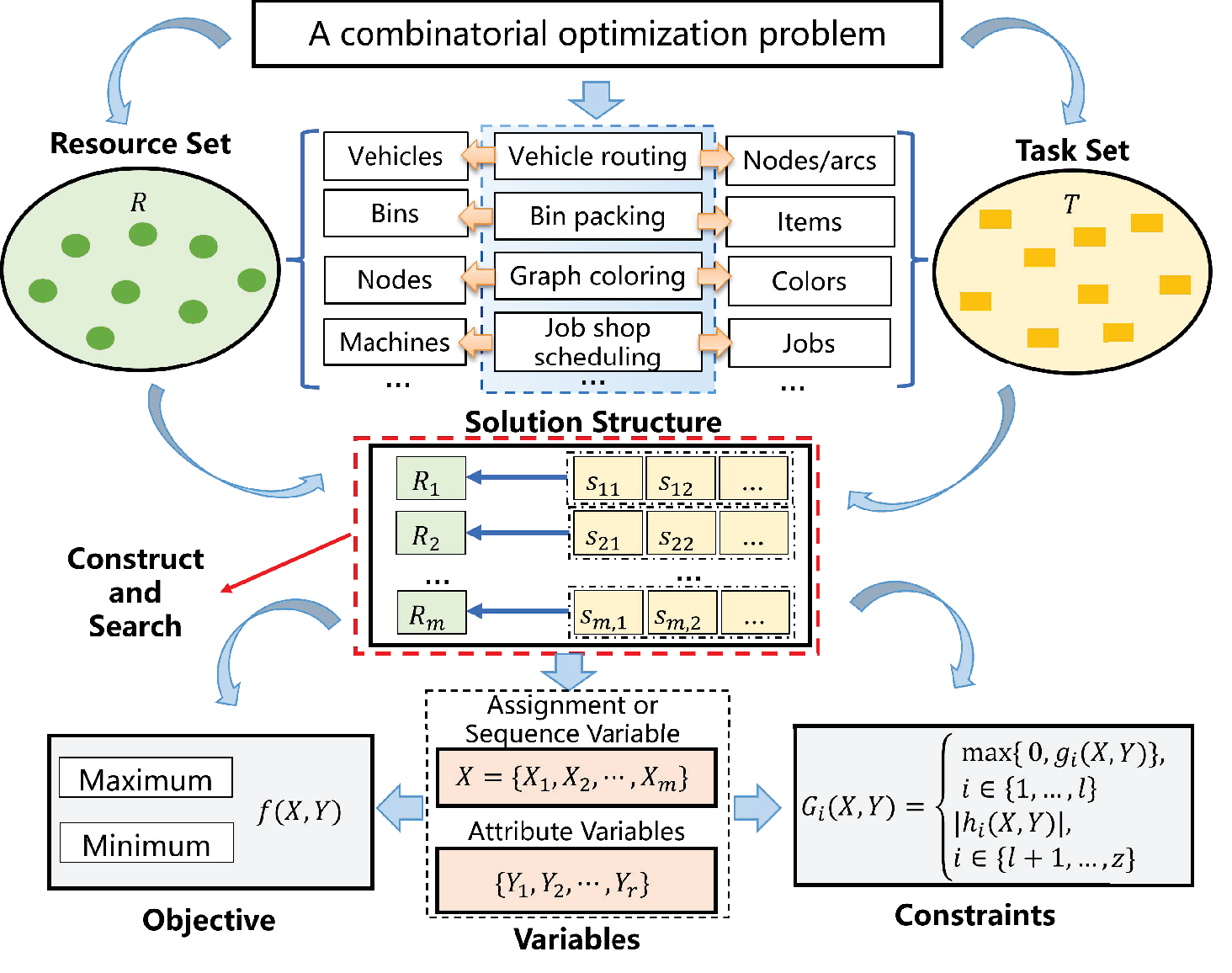, width=3.6in}}
		\end{center}
		\caption{\textcolor{black}{Modeling framework of REMS.}}\label{Fig:modeling framework}
	\end{figure}

	The modeling framework is displayed in Fig.~\ref{Fig:modeling framework}, where we first extract resources and tasks from a COP and organize them into sets $R$ and $T$, respectively. Given predetermined resources, the solution structure is represented \textcolor{black}{by} assigning tasks to specific positions of resources. Within this structure, the resources remain fixed, and the tasks from $T$ are \textcolor{black}{assigned} to each resource in $R$. We employ $S = \{ {S_1}, \cdots ,{S_m}\} $ to represent the solution structure, where $m$ is the number of resources. Specifically, $S_i=\{s_{i,1},s_{i,2},\cdots\}$ is the task set assigned to the $i$th resource, where $s_{i,k}$ indicates the task located in the $k$th position of the $i$th resource, ${s_{i,k}} \in T$.
	
	Based on the unified solution structure, we model various COPs into a unified framework by defining variables, objectives, and constraints. The variables serve to capture the specific attributes or aspects of the solution structure, including assignment or sequence variable $X$ and attribute variables $Y$. $X$ has the same form as the solution structure, explicitly representing the specific tasks assigned to resource positions. When the positions of resources are unordered, $X$ serves as an assignment variable, as in the bin packing and graph coloring problems. When the resource positions are ordered, $X$ also reflects the execution sequence of tasks. In this case, it can be regarded as a sequence variable, as seen in timetable scheduling and routing problems. The attribute variable $Y$ is used to capture other attributes of the solution structure, such as the start and end times of tasks at resource positions.
	
	The objective function $f$ represents the desired outcome of a COP based on the resource-task assignment, expressed as maximizing or minimizing a particular performance metric. The constraint $G$ reflects limitations or requirements of resources and tasks. Specifically, constraints ensure the matching feasibility among resources and tasks, such as the capacity limitations of resources and the logical relationship of the task execution sequence. The objective and constraints can be evaluated using $X$ and $Y$.
	
	Below is a detailed introduction to the key modeling elements.

	\subsubsection{\textcolor{black}{Resource and Task}}\label{s221}
	
	Assume that we can extract $m$ resources and $n$ tasks from a COP, the resource set $R$ is represented as
	\begin{equation}\label{Eq:resource}
	\begin{array}{l}
	R = \{ {R_1},{R_2}, \cdots ,{R_m}\} 
	\end{array},
	\end{equation}
	and the task set $T$ is denoted as
	\begin{equation}\label{Eq:task}
	\begin{array}{l}
	T = \{ {T_1},{T_2}, \cdots ,{T_n}\} 
	\end{array}.
	\end{equation}
	
	Each resource or task is associated with a unique index that distinguishes it from others. Additionally, both resource and task sets may possess one or more attributes, with each resource or task associated with a specific attribute value.
	
	Suppose $R$ has $p$ attributes, the index and attribute values of ${R_i}$ are expressed as
	\begin{equation}\label{Eq:resouce denote}
	\begin{array}{l}
	(d_i^R,A_{i,1}^R, \cdots ,A_{i,p}^R),{\kern 5pt}i \in \mathcal{I} = \{ 1, \cdots ,m\}   
	\end{array},
	\end{equation}
	where $d_i^R \in {\mathbb{N}}$ is the index of ${R_i}$ and $A_{i,k}^R,k \in \{ 1, \cdots ,p\}$ is the $k$th attribute value of ${R_i}$.  
	
	Likewise, the index and attribute values of ${T_j}$ with $q$ attributes are denoted as
	\begin{equation}\label{Eq:task denote}
	\begin{array}{l}
	(d_j^T,A_{j,1}^T, \cdots ,A_{j,q}^T),{\kern 5pt}j \in \mathcal{J} = \{ 1, \cdots ,n\}  
	\end{array},
	\end{equation}
	where $d_j^T \in {\mathbb{N}}$ is the index of ${T_j}$, and $A_{j,k}^T,k \in \{ 1, \cdots ,q\}$ is the $k$th attribute value of ${T_j}$. We organize all \textcolor{black}{indices} of resources and tasks into ${D^R} = \{ d_1^R, \cdots ,d_m^R\}$ and ${D^T} = \{ d_1^T, \cdots ,d_n^T\}$, respectively. 
	
	Take a VRP with $m$ vehicles and $n$ customers as an example, vehicles are treated as resources and customers as tasks, which are respectively included into $R$ and $T$. Accordingly, they can be indexed as ${D^R} = \{ 1,\cdots,m\} $ and ${D^T} = \{ 1,\cdots,n\} $. The capacity, maximum transportation distance, fixed cost, and \textcolor{black}{departure} depot for the vehicles can be regarded as the attributes of $R$. The attributes of $T$ may include the demand, the earliest and latest allowed service time, and service duration for the customers. 
	
	
	\subsubsection{Variables}\label{s223}
	
	Derived from the unified solution structure, the assignment or sequence variable can be expressed as
	\begin{equation}\label{Eq:decision variable}
	\begin{array}{l}
	X = \{ {X_1},{X_2}, \cdots ,{X_m}\} 
	\end{array},
	\end{equation}
	where ${X_i} = \{ {x_{i,1}},{x_{i,2}}, \cdots ,{x_{i,{n_i}}}\} ,i \in \mathcal{I} $ denotes the task subset assigned to resource ${R_i}$. ${n_i}$ signifies the number of available positions for ${R_i}$. ${x_{i,k}},i \in \mathcal{I},k \in {\mathcal{K}_i} = \{ 1, \cdots ,{n_i}\} $ represents the task assigned to the $k$th position of ${R_i}$. Since the index can distinguish different tasks, ${x_{i,k}} \in {D^T}$. When the positions of ${R_i}$ are \textcolor{black}{unordered}, ${X_i}$ is an unordered set and can be viewed as an assignment variable; otherwise, ${X_i}$ is an ordered sequence and regarded as a sequence variable. $X$ is consistent with the solution structure and serves as the core decision variable in a COP. Therefore, it is essential to define $X$ for any COP modeled in REMS, and the solution structure can be derived from $X$ when solving a COP.
	
	The attribute variables reflect other specific attributes of the solution structure. For example, when the solution structure is a sequence, the time-related variables, such as the start and end \textcolor{black}{service} times for tasks at specific resource positions, need to be optimized. Given that each task has a specific demand, the amount \textcolor{black}{of demand} for a task provided by a resource needs to be determined. For convenience, we first convert the variable $X$ into a binary matrix $Y^C$ and treat it as an attribute variable, where $y_{i,j,k}^C,i \in \mathcal{I},j \in \mathcal{J},k \in {\mathcal{K}_i}$ indicates whether $T_j$ is assigned to the $k$th position of $R_i$ and is obtained by
	\begin{equation}\label{Eq:0-1 decision variable}
	\begin{array}{l}
	y_{i,j,k}^C = \left\{ {\begin{array}{*{20}{l}}
		{1,{\kern 5pt}{\rm{ if }}{\kern 5pt}{x_{i,k}} = d_j^T}\\
		{0,{\kern 5pt}{\rm{ otherwise}}}
		\end{array}} \right.
	\end{array}.
	\end{equation}
	
	Accordingly, for the above-mentioned VRP, we use ${t_{{j},{j'}}},{j},{j'} \in \{ 0, \cdots ,n \} $ to represent the travel time from ${j}$ to ${j'}$, where $0$ denotes the index of the depot and  $\{ 1, \cdots ,n\} $ corresponds to the index of tasks. An attribute variable $Y^S$ is employed to represent the start \textcolor{black}{service} times, where $y_{i,j,k}^S, i \in \mathcal{I},j \in \mathcal{J},k \in {\mathcal{K}_i}$ denotes the start \textcolor{black}{service} time for ${T_j}$ in the $k$th position of ${R_i}$ and can be calculated by
	\begin{equation}\label{Eq:start serve time}
	\begin{array}{l}
	y_{i,j,k}^S = \left\{ {\begin{array}{*{20}{l}}
		{{t_{0,j}}y_{i,j,k}^C,{\kern 5pt}k = 1}\\
		{\sum\limits_{j' \in \mathcal{J}} {(y_{i,j',k - 1}^S + {t_{j',j}})} y_{i,j,k}^C,{\kern 5pt}k > 1}
		\end{array}} \right..
	\end{array}
	\end{equation}
	
	
	As task ${T_j}$ has a \textcolor{black}{service} duration ${s_j}$, the end \textcolor{black}{service} time $y_{i,k}^E$ for $T_j$ located in the $k$th position of $R_i$ is calculated by
	\begin{equation}\label{Eq:end serve time}
	\begin{array}{l}
	y_{i,j,k}^E = (y_{i,j,k}^S + {s_j})y_{i,j,k}^C,{\kern 5pt}i \in \mathcal{I},k \in {\mathcal{K}_i}
	\end{array}.
	\end{equation}
	
	When task ${T_j}$ has a demand of ${q_j}$, we use another attribute variable $y_{i,j,k}^Q,i \in \mathcal{I},k \in {\mathcal{K}_i}$ to indicate the amount of demand provided to $T_j$ in the $k$th position of ${R_i}$. It is calculated by
	\begin{equation}\label{Eq:serve amount}
	\begin{array}{l}
	y_{i,j,k}^Q = {q_j}y_{i,j,k}^C,{\kern 5pt}i \in \mathcal{I},k \in {\mathcal{K}_i}
	\end{array}.
	\end{equation}
	
	As seen in \eqref{Eq:0-1 decision variable}-\eqref{Eq:serve amount}, the attribute variables can be expressed as a function of $X$. If we define $r$ attribute variables for a COP, the $i$th attribute variable can be expressed as
	 \begin{equation}\label{Eq:state variable}
	 \begin{array}{l}
	 {Y_i} = {\psi _i}(X),{\kern 5pt}i \in \{ 1, \cdots ,r\} 
	 \end{array}.
	 \end{equation}
	 
	
	\subsubsection{Objectives}\label{s224}
	
	The objective function varies with the problem type and specific requirements. Some frequently occurring objective functions in COPs include minimizing the total cost, maximizing the profit, and minimizing the makespan.
	
	For example, given that assigning $T_j$ to the $k$th position of ${R_i}$ has a cost ${c_{i,j,k}}$, the objective of minimizing the total cost can be expressed as
	\begin{equation}\label{Eq:total cost of resources}
	\begin{array}{l}
	{\rm{min  }}\sum\limits_{k \in {K_i}} {\sum\limits_{j \in J} {\sum\limits_{i \in I} {{c_{i,j,k}}y_{i,j,k}^C} } } 
	\end{array}
	\end{equation}
	
	Additionally, assume that putting task ${T_j}$ to the $k$th position of resource $R_i$ generates profit ${p_{i,j,k}}$, and then maximizing the profit can be represented as
	\begin{equation}\label{Eq:maximize the profit}
	\begin{array}{l}
	{\rm{max  }}\sum\limits_{i \in \mathcal{I}} {\sum\limits_{j \in \mathcal{J}} {\sum\limits_{k \in {\mathcal{K}_i}} {({p_{i,j,k}} - {c_{i,j,k}})y_{i,j,k}^C} } } 
	\end{array}
	\end{equation}
	
	
	For minimizing the makespan, we take the above-mentioned VRP as an example. Therefore, the objective is denoted as
	\begin{equation}\label{Eq:minimize the makespan}
	\begin{array}{l}
	\min \left\{ \max_{i \in \mathcal{I}, j \in \mathcal{J}, k \in \mathcal{K}_i} y_{i,j,k}^E \right\}
	\end{array}
	\end{equation}
	
	Overall, as shown in the examples of \eqref{Eq:total cost of resources}-\eqref{Eq:minimize the makespan}, the objective functions can be calculated \textcolor{black}{from} variables $X$ and $Y$, i.e.,
	\begin{equation}\label{Eq:objective}
	{\rm{min {\kern 5pt} or {\kern 5pt} max {\kern 5pt}}}f(X,Y)
	\end{equation}
	
	\subsubsection{Constraints}\label{s225}
	
	The constraints reflect limitations imposed by resources and tasks, encompassing logical relationships, the ranges of variables, resource capacities, task-related requirements on time or space, etc.. The following are some common constraints.
	
	There is a class of constraints related to resources, focusing on their capacities, availability, or limitations. Such constraints impose a limitation on the number of tasks assigned to each resource, which can be formulated as
	\begin{equation}\label{Eq:resource constraint 1}
	\begin{array}{l}
	\sum\limits_{k \in {\mathcal{K}_i}} {\sum\limits_{j \in \mathcal{J}} {y_{i,j,k}^C} } ( \le , \ge {\rm{or}} = )n_i^R,{\kern 5pt}i \in \mathcal{I}
	\end{array},
	\end{equation}
	where $n_i^R$ is the threshold value for the number of tasks assigned to ${R_i}$. The constraint in \eqref{Eq:resource constraint 1} ensures that the number of tasks assigned to ${R_i}$ is no more than, no less than, or exactly equal to $n_i^R$, depending on the specific requirement.
	
	Moreover, if the capacity of resource ${R_i}$ is ${Q_i}$ and the demand for task ${T_j}$ is ${q_j}$, the capacity constraint \textcolor{black}{on} resources can be defined as
	\begin{equation}\label{Eq:resource constraint 2}
	\begin{array}{l}
	\sum\limits_{k \in {\mathcal{K}_i}} {\sum\limits_{j \in \mathcal{J}} {{q_j}y_{i,j,k}^C} }  \le {Q_i},{\kern 5pt}i \in \mathcal{I}
	\end{array}.
	\end{equation}
	
	Furthermore, the constraints presented in \eqref{Eq:resource constraint 1} and \eqref{Eq:resource constraint 2} can be formulated in a unified form as
	\begin{equation}\label{Eq:resource constraint 3}
	\begin{array}{l}
	\sum\limits_{k \in {\mathcal{K}_i}} {\sum\limits_{j \in \mathcal{J}} {{\rho _{i,j,k}^R}y_{i,j,k}^C} } ( \le , \ge {\rm{or}} = )\theta_i^R,{\kern 5pt}i \in \mathcal{I}
	\end{array},
	\end{equation}
	where $\theta_i^R$ is a threshold value. When ${\rho _{i,j,k}^R} = 1$ and $\theta_i^R = n_i^R$, the constraint is \textcolor{black}{reduced} to \eqref{Eq:resource constraint 1}. When ${\rho _{i,j,k}^R} = {q_j},i \in \mathcal{I},k \in {\mathcal{K}_i}$ and $\theta_i^R = {Q_i}$, it becomes constraint \eqref{Eq:resource constraint 2}.
	
	Additionally, some constraints focus on task-related limitations. For instance, a constraint specifying that task ${T_j}$ is executed no more than, no less than, or exactly $n_j^T$ times on all resources $R$ can be denoted as
	\begin{equation}\label{Eq:task constraint 1}
	\begin{array}{l}
	\sum\limits_{i \in \mathcal{I}} {\sum\limits_{k \in {\mathcal{K}_i}} {y_{i,j,k}^C} } ( \le , \ge {\rm{or}} = )n_j^T,{\kern 5pt}j \in \mathcal{J}
	\end{array}.
	\end{equation}
	
	The constraints like \eqref{Eq:task constraint 1} can further unified as:
	\begin{equation}\label{Eq:task constraint 2}
	\begin{array}{l}
	\sum\limits_{k \in {\mathcal{K}_i}} {\sum\limits_{i \in \mathcal{I}} {{\rho _{i,j,k}^T}y_{i,j,k}^C} } ( \le , \ge {\rm{or}} = )\theta_i^T,{\kern 5pt}j \in \mathcal{J}
	\end{array},
	\end{equation}
	where $\theta_i^T$ indicates the threshold value. When $\theta_i^T = n_j^T$ and ${\rho _{i,j,k}^T} = 1$, it can be reduced to constraint \eqref{Eq:task constraint 1}. 
	
	When it comes to \textcolor{black}{modeling} the logical relationships among tasks, it may be necessary to specify whether tasks ${T_{{j}}}$ and ${T_{{j'}}}$ must be assigned to the same resource or to different ones. Such constraints can be expressed as
	\begin{equation}\label{Eq:logical constraint 1}
	\begin{array}{l}
	\sum\limits_{k \in {{\cal K}_i}} {y_{i,j',k}^C}  = 0 {\kern 5pt}{\rm{ or }}{\kern 5pt} 1,{\kern 5pt}{\rm{ if }}\sum\limits_{k \in {{\cal K}_i}} {y_{i,j,k}^C}  = 1,j \ne j' \in {\cal J},i \in {\cal I}
	\end{array}.
	\end{equation}
	
	Some COPs impose constraints on \textcolor{black}{the execution order of tasks}. If task ${T_{{j'}}}$ must be arranged before ${T_{{j}}}$, it is denoted as
	\begin{equation}\label{Eq:logical constraint 2}
	\begin{array}{r}
	\sum\limits_{k' \in \{ 1, \cdots ,k - 1\} } {y_{i,j,k}^C}  = 1,{\rm{ {\kern 5pt} if {\kern 5pt} }}{x_{i,k}} = d_{{j'}}^T,i \in \mathcal{I},\\
	{j} \ne {j'} \in \mathcal{J},k \in {\mathcal{K}_i}
	\end{array}.
	\end{equation}
	
	Another class of constraints defines the relationships between resources and tasks. For example, the constraint that task ${T_j}$ needs to be executed on resource ${R_i}$ at most, at least, or exactly $n_{i,j}^T$ can be
	\begin{equation}\label{Eq:resource-task constraint 1}
	\begin{array}{l}
	\sum\limits_{k \in {{\cal K}_i}} {y_{i,j,k}^C} ( \le , \ge {\rm{or}} = )n_{i,j}^T,{\kern 5pt}i \in \mathcal{I},j \in \mathcal{J}
	\end{array}.
	\end{equation}
	
	The constraints, such as \eqref{Eq:resource-task constraint 1}, can be further unified as
	\begin{equation}\label{Eq:resource-task constraint 2}
	\begin{array}{l}
	\sum\limits_{k \in {{\cal K}_i}} {{\rho_{i,j,k}}y_{i,j,k}^C} ( \le , \ge {\rm{or}} = )\theta_{i,j},{\kern 5pt}i \in \mathcal{I},j \in \mathcal{J}
	\end{array},
	\end{equation}
	where $\theta_{i,j}$ is the threshold value and ${\rho_{i,j,k}}$ denotes the coefficient. If $\theta_{i,j}{\rm{ = }}n_{i,j}^T$ and ${\rho_{i,j,k}}{\rm{ = }}1$, \eqref{Eq:resource-task constraint 2} is converted into \eqref{Eq:resource-task constraint 1}.
	
	Similar to the objective function, constraints can also be expressed in terms of variables $X$ and $Y$, as shown in \eqref{Eq:resource constraint 1}-\eqref{Eq:resource-task constraint 2}. These constraints are categorized into equality and inequality constraints and are commonly expressed as \eqref{Eq:equality and inequality constraints}.
	\begin{equation}\label{Eq:equality and inequality constraints}
	\begin{array}{l}
	\left\{ {\begin{array}{*{20}{l}}
		{{g_i}(X,Y) \le 0,{\kern 5pt}i \in \{ 1, \cdots ,l\} }\\
		{{h_i}(X,Y) = 0,{\kern 5pt}i \in \{ l + 1, \cdots ,z\} }
		\end{array}} \right.
	\end{array},
	\end{equation}
	where ${g}(X,Y)$ and ${h}(X,Y)$ denote the inequality and equality constraints, respectively.
	
	To evaluate the $i$th constraint value for a solution, it is calculated by
	\begin{equation}\label{Eq:constraints}
	\begin{array}{l}
	{G_i}(X,Y) = \left\{ {\begin{array}{*{20}{l}}
		{\max \{ 0,{g_i}(X,Y)\} ,{\kern 5pt}i \in \{ 1, \ldots ,l\} }\\
		{|{h_i}(X,Y)|,{\kern 5pt}i \in \{ l + 1 \ldots ,z\} }
		\end{array}} \right.
	\end{array}.
	\end{equation}

	Overall, given the precondition that resources and tasks remain discrete and unchanged, we propose a unified solution structure for the general COPs. Based on this, the assignment or sequence variable is expressed in a standardized form, as seen in \eqref{Eq:decision variable}. The attribute variables can be freely defined \textcolor{black}{without restrictions on their type or structure}, as long as they can be obtained from $X$. Additionally, we do not impose any specific requirements on the objective and constraint functions. Instead, we require only that these functions provide a quantifiable value to evaluate solutions.

	\section{Design of metaheuristic algorithms}\label{s3}

	\subsection{\textcolor{black}{Design Framework of Metaheuristics in REMS}}\label{s31}

	\textcolor{black}{When applying metaheuristics to solve COPs, the specific problem knowledge related to variables, objectives, and constraints is comprehensively considered to encode solutions and design problem-specific operators. Specifically, various solutions are first encoded in various solution structures, such as sequences \cite{lau2009application} and matrices \cite{saviniec2017effective}. The solution structure defines the search space and guides the design of algorithms. Subsequently, based on the solution structure and the specific objectives and constraints, diverse problem-specific operators are designed to construct a solution and search the solution space \cite{lau2009application, zhen2009hybrid}. When infeasible solutions arise, the constraint handling methods, such as the penalty function \cite{sun2023hybrid, saviniec2017effective} and repair operators \cite{lau2009application}, are developed to guide the search toward feasibility. However, due to the limitation of specific problem domain knowledge, any variation in a COP may render the original algorithm inefficient or even inapplicable.}
	
\begin{figure}[htb]
		\begin{center}		
			\subfigure{\psfig{file=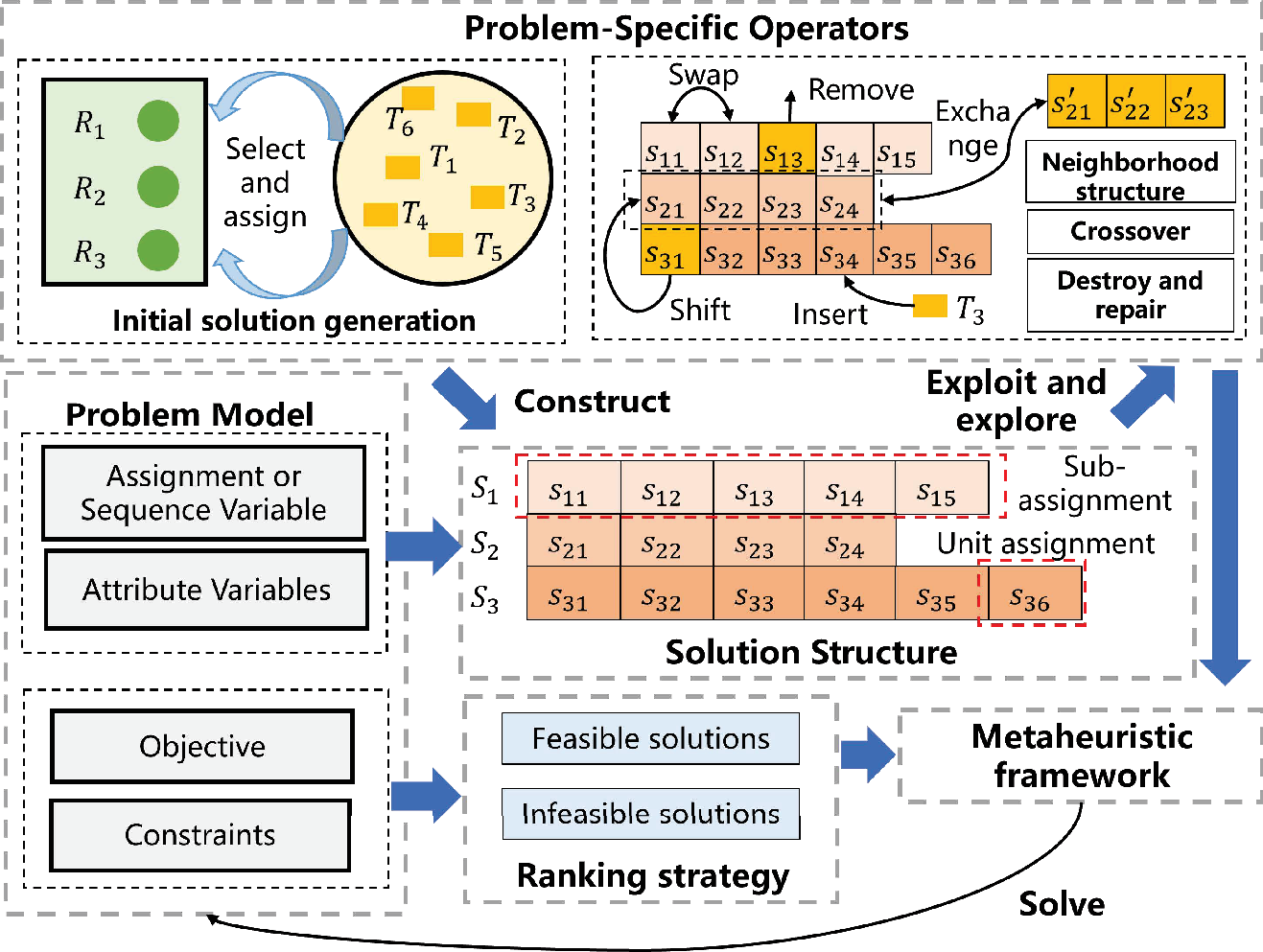, width=3.4in}}
		\end{center}
		\caption{\textcolor{black}{Design framework for metaheuristic algorithms in REMS.}}\label{Fig:solution structure}
	\end{figure}

	\textcolor{black}{The design principle and logic of the operators and algorithms proposed in the REMS framework are similar to those adopted in most existing studies. Moreover, since REMS extracts and utilizes more general problem domain knowledge from various COPs, the solution structure and problem-specific operators are accordingly different from previous research. Fig.~\ref{Fig:solution structure} provides the design framework for metaheuristic algorithms used to solve the problem modeled in REMS.}
	
	\textcolor{black}{The solution structure $S$ is first unified as the assignment of tasks to resource positions. For convenience, assigning a task to a specific position is referred to as a unit assignment. All unit assignments for a resource form a sub-assignment, and the sub-assignments for all resources constitute a solution. Corresponding to the assignment and sequence variable, the sub-assignment is either an ordered sequence of tasks (for sequence variables) or an unordered set of tasks (for assignment variables). Subsequently, during the optimization, we construct the unified solution structure and search the solution space from the resource-task-oriented perspective. Specifically, an initial strategy is designed to construct $S$ through the step-by-step insertion of feasible tasks into resource positions. The solution space is exploited and explored by a series of changes, including shifting, swapping, removing, exchanging, and inserting tasks across multiple resources or within a single resource. Based on the unified formulations of objectives and constraints, a hierarchical ranking method is developed that accounts for the diverse scales of objectives and constraints. The proposed operators are integrated into different metaheuristic frameworks, forming various metaheuristic methods to solve the problems modeled in REMS. }

	\subsection{Initial Solution}\label{s32}
	
\begin{figure}[htb]
		\begin{center}		
			\subfigure{\psfig{file=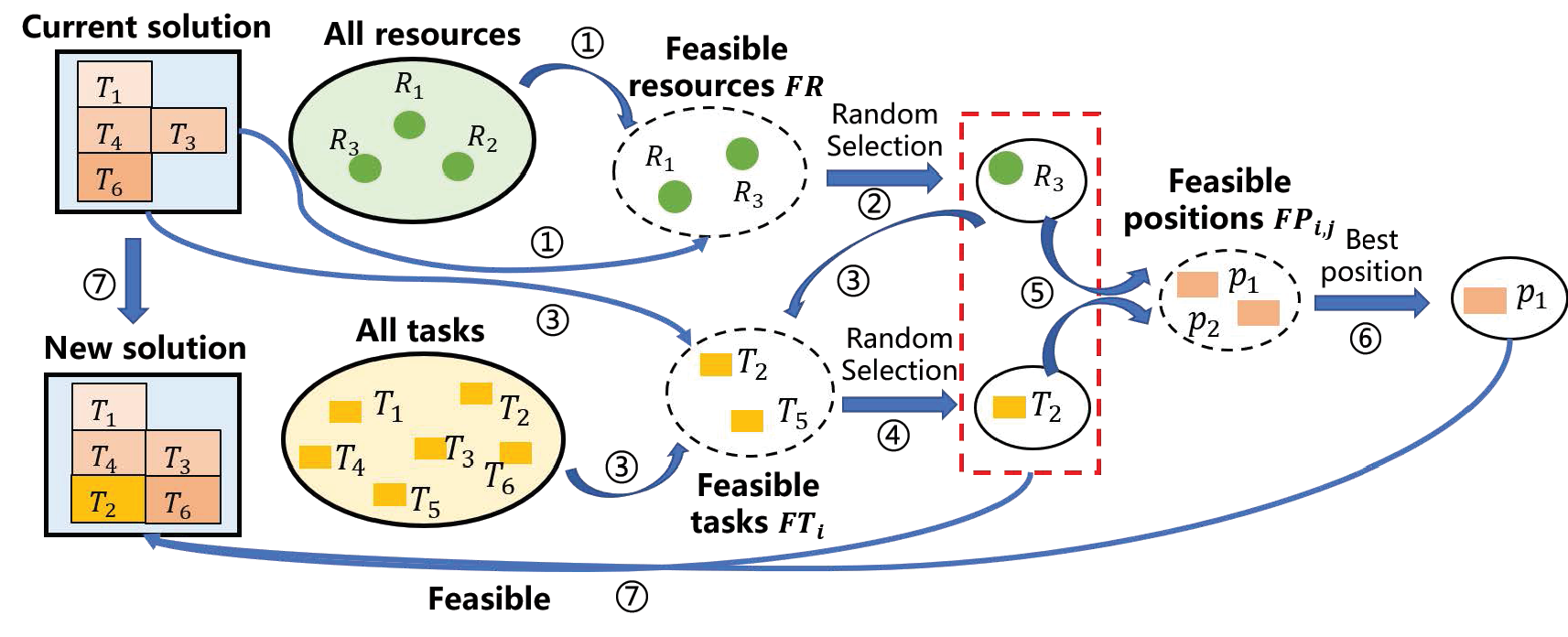, width=3.6in}}
		\end{center}
		\caption{Determine a feasible unit assignment for a current solution.}\label{Fig:feasible assigment}
	\end{figure}

	An initial solution is generated by a constructive heuristic rule, where unit assignments are incrementally determined starting from an empty solution. To ensure feasibility, each unit assignment should be inserted without violating constraints until no further feasible unit assignments can be inserted. Determining the feasible unit assignment, i.e., feasible resources, tasks, and positions, in each step is the core of constructing an initial solution. To avoid searching \textcolor{black}{for} infeasible unit assignments, we preliminarily determine the feasible unit assignment using some common constraints information presented in Section~\ref{s225}. Determining a feasible unit assignment for a current solution is displayed in Fig.~\ref{Fig:feasible assigment}.
	
	
	First, a feasible resource set ${FR}$ is identified from the current solution according to constraint \eqref{Eq:resource constraint 3}. Following this, a resource ${R_i}$ is randomly selected from ${FR}$. Subsequently, according to the constraints in \eqref{Eq:task constraint 2}, \eqref{Eq:logical constraint 1}, and \eqref{Eq:resource-task constraint 2}, we determine the feasible task set $FT_i$ for ${R_i}$. A task ${T_j}$ is then randomly selected from $FT_i$. When ${S_i}$ is an unordered set, the insertable position is ${A} = \{ |{S_i}| + 1\} $. Otherwise, if ${S_i}$ is an ordered sequence, the insertable positions are ${A} = \{ 1, \cdots ,|{S_i}| + 1\} $. Based on constraints in \eqref{Eq:logical constraint 2}, the feasible positions $FP_{i,j} \subseteq {A}$ are further determined. Subsequently, the best position ${p_k}$, \textcolor{black}{which} enables the best solution among $FP_{i,j}$, is selected. If the unit assignment of assigning ${T_j}$ to position ${p_k}$ of ${R_i}$ enables an equal or better solution, then the current solution is updated by inserting the unit assignment.
	
	Algorithm~\ref{PC:feasible assignment} outlines the process for determining a unit assignment for the current solution $S$. Before generating an initial solution, the infeasible resource set $inR$ and the infeasible task sets $inT_i,i \in \mathcal{I}$ are initially empty. Lines 1-2 select a resource ${R_i}$ from ${FR}$, and lines 3-4 randomly select a task ${T_j}$ from $FT_i$. Lines 5 assign ${T_j}$ to the best position of ${R_i}$, thereby forming ${S_{\rm{candi}}}$. In lines 6-13, if ${S_{\rm{candi}}}$ is no worse than $S$, the new unit assignment is accepted. Accordingly, $inR$ and $inT_i,i \in \mathcal{I}$ are updated.
	\begin{algorithm}[htp]
		\caption{\textbf{Function} \emph{FeasibleAssignment}()}\label{PC:feasible assignment}
		\KwIn{Problem model, current solution $S$, $in{R}$, and $inT_i,i \in \mathcal{I}$.}
		\KwOut{${S_{\rm{new}}}$, $inR$, and $inT_i,i \in \mathcal{I}$.}
		\LinesNumbered
		Identify ${FR}$ for ${S}$ and ${FR} \leftarrow {FR}\backslash in{R}$ \;
		Randomly select a resource ${R_i}$ from ${FR}$\;
		Determine $FT_i$ and $FT_i \leftarrow FT_i\backslash inT_i$\;
		Randomly select a task ${T_j}$ from $FT_i$\;
		Insert the ${T_j}$ into the best position of ${R_i}$ for ${S}$ and form ${S_{\rm{candi}}}$\;
		\If{${S_{\rm{candi}}}$ is no worse than ${S}$}{
			${S_{\rm{new}}} \leftarrow {S_{\rm{candi}}}$\;
			\If{${S_i}$ is a sequence}{
				$inT_i \leftarrow \emptyset $\;
			}
			}
		\Else{
			${S_{\rm{new}}} \leftarrow S$, and $inT_i \leftarrow inT_i \cup {T_j}$ \;
			\If{$T\backslash inT_i = \emptyset $}{
				$in{R} \leftarrow in{R} \cup {R_i}$\;
			}
		}		
	\end{algorithm}
	
	Algorithm~\ref{PC:Initial Solution} displays the procedure for generating an initial solution, where tasks are iteratively assigned to available positions on resources until no feasible task can be inserted.
	\begin{algorithm}[htp]
		\caption{\textbf{Function} \emph{InitialSolution}()}\label{PC:Initial Solution}
		\KwIn{Problem model.}
		\KwOut{Initial solution $S$.}
		\LinesNumbered
		Initialize ${S_i} \leftarrow \emptyset ,i \in \mathcal{I}$, $in{R} \leftarrow \emptyset $, and $inT_i \leftarrow \emptyset ,i \in \mathcal{I}$\;
		\While{$R\backslash inR \ne \emptyset $}{
			$[S,inR,inT_i,i \in \mathcal{I}] \leftarrow $\emph{FeasibleAssignment}()\;
		}
	\end{algorithm}
	
	\subsection{Neighborhood Structure}\label{s33}
	
	The neighborhood \textcolor{black}{structure} is employed to perturb the current solution, which serves as a fundamental component in the neighborhood-based algorithms, such as VNS, SA, and TS. To efficiently solve the COPs in REMS, we propose the following six types of neighborhood structures \textcolor{black}{from the resource-task perspective} based on the unified solution structure.
	
	\begin{itemize}
		\item \emph{Swap}: It selects two sets of tasks, each containing 1-2 consecutive tasks from either the same resource or two different resources, and exchanges them.
		
		\item \emph{Shift}: It first selects a set of tasks from a resource and then shifts it to other positions within the same resource or different resources. To facilitate a diverse neighborhood structure, when shifting within the same resource, the selected task set contains 1-2 consecutive tasks. When shifting across different resources, the selected task set may consist of either 1-2 consecutive tasks or all tasks from the resource. 
		
		\item \emph{Remove}: The operation selects a task and removes it from the current solution.
		
		\item \emph{Insert}: It selects a task and inserts it into a feasible position of the solution.
		
		\item \emph{Remove and insert}: It selects a set with 1-2 consecutive tasks to remove. Subsequently, other feasible tasks are inserted into the current solution.
		
		\item \emph{Reverse}: It selects a set of tasks within a resource or spanning two resources and reverses the order of these tasks. The reversed tasks are then reinserted according to their new order.
		
	\end{itemize}
	
	Among these neighborhood structures, the insert operation should be performed without increasing any constraint value to ensure the feasibility of the neighborhood solution.
	
	\textcolor{black}{Based on the six types of neighborhood structures, various specific neighborhood structures can be derived by adjusting the number of changed tasks and whether these changes occur within the same or across different resources. The effectiveness of neighborhood structures is influenced by the problem characteristics, current solution, and historical search information. To adaptively select appropriate neighborhood structures by exploring the available information, a Q-learning-based selection strategy is proposed.}
	
	\textcolor{black}{In the strategy, $ns$ operators are regarded as actions, forming the action set ${\mathcal{A}} = \{ O{P_1}, \cdots ,O{P_{ns}}\} $. The state denoted by $\vec s$ includes the information from the problem level, solution level, and search process.}
	
	\textcolor{black}{The problem-level states capture information such as the number of resources and variable types. Let ${s_1} = 1$ when the number of resources equals 1; otherwise, ${s_1} = 2$. ${s_2} = 1$ if $X$ is an assignment variable; otherwise, ${s_2} = 2$. The solution-level states involve constraint-related information and the maximum number of tasks assigned across all resources. Specifically, we have ${s_3} = 1$ if the solution is feasible and ${s_3} = 2$ if it is infeasible. We set ${s_4} = 1$, ${s_5} = 1$, ${s_6} = 1$, ${s_7} = 1$, and ${s_8} = 1$ if the constraints in \eqref{Eq:resource constraint 3}, \eqref{Eq:task constraint 2}, \eqref{Eq:logical constraint 1}, \eqref{Eq:logical constraint 2} and \eqref{Eq:resource-task constraint 2} are satisfied, respectively. Otherwise, the corresponding state is set to 2. If the maximum number of tasks across all resources for the solution does not exceed 1, we have ${s_9} = 1$. When the maximum equals 2, ${s_9} = 2$; and when it exceeds 2, ${s_9} = 3$. The search process state captures whether stagnation occurs. If the current solution fails to improve after several consecutive applications of neighborhood structures, ${s_{10}} = 1$; otherwise, ${s_{10}} = 2$.}
	
	\textcolor{black}{In the $t$th step, the selected action and state of the current solution ${S_t}$ are separately denoted as ${a_t} \in {\mathcal{A}}$ and $\vec s_t$. Executing ${a_t}$ for ${S_t}$ generates a new solution ${S_{t + 1}}$. The reward function for executing ${a_t}$ under the state $\vec s_t$ is defined as}
	\begin{equation}\label{Eq: reward function}
	\color{black}{{r_t}({\vec s_t},{a_t}) = \left\{ {\begin{array}{*{20}{l}}
			{2,{\rm{ {\kern 4pt}if {\kern 4pt}}}{S_{t + 1}}{\rm{{\kern 4pt} is {\kern 4pt}better{\kern 4pt} than{\kern 4pt} }}{S_t}}\\
			{1,{\rm{ {\kern 4pt}if{\kern 4pt} }}{S_{t + 1}}{\rm{{\kern 4pt} is{\kern 4pt} worse{\kern 4pt} than{\kern 4pt} }}{S_t}{\rm{{\kern 4pt} in{\kern 4pt} objectives}}}\\
			{0,{\kern 4pt}{\rm{ if }}{\kern 4pt}{S_{t + 1}}{\rm{ {\kern 4pt}is {\kern 4pt}equal{\kern 4pt} to{\kern 4pt} }}{S_t}}\\
			{ - 1,{\kern 3pt}{\rm{ if {\kern 3pt}}}{S_{t + 1}}{\rm{ {\kern 3pt}is{\kern 3pt} worse{\kern 3pt} than{\kern 3pt} }}{S_t}{\rm{{\kern 3pt} in{\kern 3pt} contraints}}}\\
			{ - 2,{\kern 4pt}{\rm{ if }}{\kern 4pt}{S_{t + 1}}{\rm{{\kern 4pt} is{\kern 4pt} not{\kern 4pt} reachable}}}
			\end{array}} \right.}
	\end{equation}
	
	\textcolor{black}{As shown in \eqref{Eq: reward function}, a larger reward is assigned for executing an action in the current state that results in an improved solution. If the new solution has a worse objective value and a comparable level of constraint violations to the current one, a smaller positive reward is given to encourage exploration. When executing the action under the current state leaves the solution unchanged, a reward of zero is assigned. If the action results in a solution with larger constraint violations, a negative reward is applied. Moreover, if the action is unsuitable for the current solution, for example, changes in multiple resources but the COP contains only a single resource, the new solution is not reachable and a larger negative reward is imposed. In such cases, the states are set to 0.}
		
	\textcolor{black}{Subsequently, the Q values are updated by}
	\begin{equation}\label{Eq: Q update}
	\color{black}{\begin{array}{l}
		Q({\vec s_t},{a_t}) = Q({\vec s_t},{a_t}) + \\\alpha ({r_t}({\vec s_t},{a_t}) + \gamma \mathop {\max }\limits_{a'} Q({\vec s_{t+1}},a') - Q({\vec s_t},{a_t}))
		\end{array}},
	\end{equation}
	\textcolor{black}{where $\alpha$ is the learning rate and $\gamma$ is the discount factor. }
	
	\textcolor{black}{To balance exploration and exploitation, an improved $\varepsilon $-greedy strategy is employed. Specifically, with a greedy probability $\varepsilon $, an operator is randomly selected from $\mathcal{A}$. Otherwise, the top 20\% of actions with the largest Q values are identified, and the final action is chosen by a roulette wheel selection based on their historical success rates to further utilize the historical search information. In the roulette wheel selection, the selection probability for the $i$th neighborhood structure can be denoted as $p_i,i \in \{ 1, \cdots, {ns} \} $. $p_i$ depends on the success times $sc_i$ and selected times $st_i$ over a specific consecutive iteration, where ${p_i} = s{c_i}/s{t_i}$.}

	
	\textcolor{black}{Accordingly, Algorithm~\ref{PC:Neighborhood Solution} displays the pseudocode for generating a neighborhood solution.}
	\begin{algorithm}[htp]
		\color{black}{\caption{\textcolor{black}{\textbf{Function} \emph{NeighborhoodSolution}()}}\label{PC:Neighborhood Solution}
			\color{black}{\KwIn{Problem model, current solution $S$.}
				\KwOut{Neighborhood solution ${S_{\rm{neighbor}}}$.}
				\LinesNumbered
				\If{Q table is not built}{
					$Q \leftarrow 0$ and $p_i \leftarrow 1,sc_i, st_i \leftarrow 0,i \in \{ 1, \cdots, {ns} \} $\;
				}
				Obtain the state of $S$\;
				Select a neighborhood structure $i$ by the improved $\epsilon$-greedy strategy\;
				Generate a neighborhood solution $S_{\rm{neighbor}}$ of $S$ by the selected neighborhood structure\;
				Obtain the state of $S_{\rm{neighbor}}$ and calculate the reward value by \eqref{Eq: reward function} \;
				Update $Q$ by \eqref{Eq: Q update}, and $p_i,sc_i, st_i,i \in \{ 1, \cdots, {ns} \} $\;}}
	\end{algorithm}
	
	
	\subsection{Destroy-and-Repair Operator}\label{s34}
	
	The destroy operator randomly selects $nd$ tasks from the current solution $S$ and removes them, forming a destroy solution ${S_{\rm{destroy}}}$. Subsequently, the repair operator iteratively reinserts feasible tasks into the best positions of ${S_{\rm{destroy}}}$, \textcolor{black}{resulting in} a repair solution ${S_{\rm{repair}}}$. The destroy-and-repair procedure is outlined in Algorithm~\ref{PC:Destroy Repair}. Notably, the process of generating a repair solution is similar to that in Algorithm~\ref{PC:Initial Solution}, with the current solution being treated as ${S_{\rm{repair}}}$.
	\begin{algorithm}[htp]
		\caption{\textbf{Function} \emph{DestroyRepair}()}\label{PC:Destroy Repair}
		\KwIn{Problem model, $S$.}
		\KwOut{${S_{\rm{repair}}}$.}
		\LinesNumbered
		Randomly remove ${nd}$ tasks from $S$, the resulting solution is recorded as ${S_{\rm{destroy}}}$\;
		${S_{\rm{repair}}} \leftarrow {S_{\rm{destroy}}}$, $inR \leftarrow \emptyset $, and $inT_i \leftarrow \emptyset ,i \in \mathcal{I}$ \;
		\While{$R\backslash inR \ne \emptyset $}{
			$[{S_{\rm{repair}}},inR,inT_i,i \in \mathcal{I}] \leftarrow $ \emph{FeasibleAssignment}()\;
		}
	\end{algorithm}
	
	\subsection{Crossover Operator}\label{s35}
	
\begin{figure}[htb]
		\begin{center}	\subfigure{\psfig{file=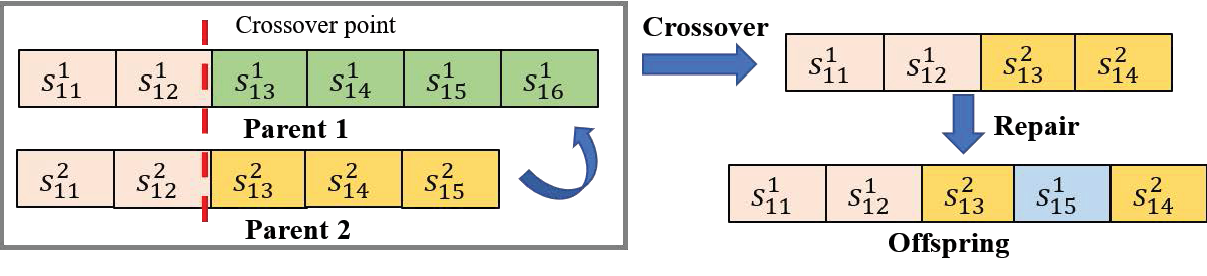, width=3.4in}}
		\end{center}
		\caption{Single-point crossover for COPs with a single resource.}\label{Fig:Single-point crossover}
	\end{figure}

	A single-point crossover operator is designed for COPs with a single resource, while a two-point crossover is applied for those involving multiple resources. As presented in Fig.~\ref{Fig:Single-point crossover}, the single-point crossover randomly selects a crossover point for Parent 1 and Parent 2. The offspring inherits the unit assignments from Parent 1 prior to the crossover point. The tasks from Parent 2 after the crossover point are sequentially added, ensuring that no constraints are violated. Lastly, the repair operation outlined in Algorithm~\ref{PC:Destroy Repair} is applied to repair the infeasible solution.
	
\begin{figure}[htb]
		\begin{center}		\subfigure{\psfig{file=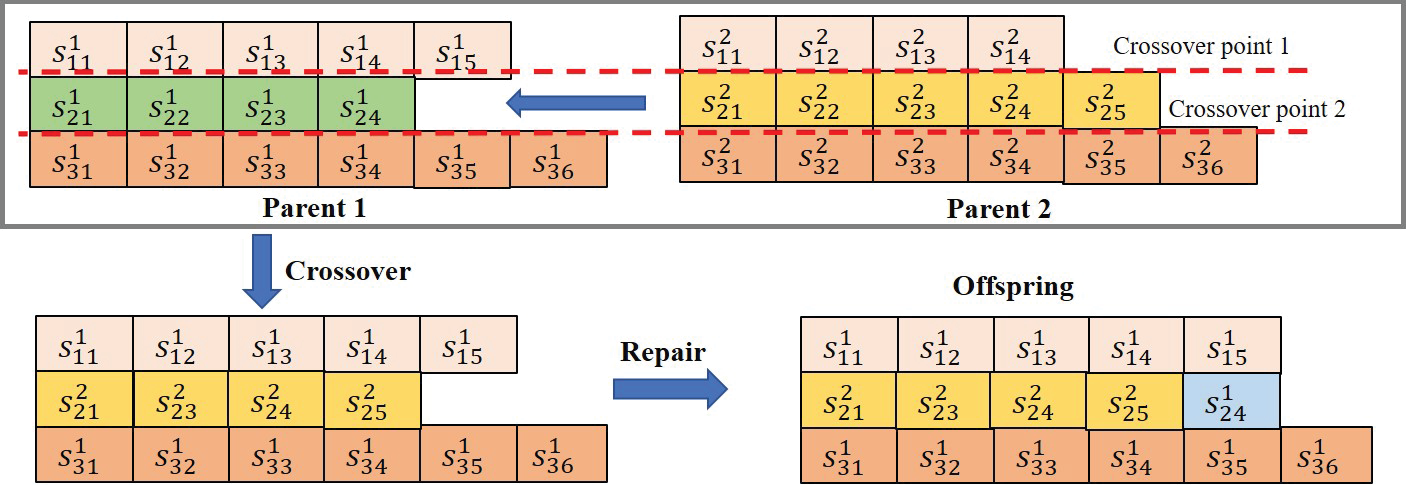, width=3.6in}}
		\end{center}
		\caption{Two-point crossover for COPs with multiple resources.}\label{Fig:Two-point crossover}
	\end{figure}

	Fig.~\ref{Fig:Two-point crossover} illustrates the process of generating offspring via a two-point crossover. First, two crossover points are randomly selected from Parents 1 and 2. Second, the offspring inherits the sub-assignments outside the two crossover points from Parent 1. The sub-assignments between the two points are inherited from Parent 2, where each unit assignment is added to the offspring without violating any constraints. Similar to the single-point crossover, the repair operation in Algorithm~\ref{PC:Destroy Repair} is employed to refine the infeasible solution.
	
	Algorithm~\ref{PC:Crossover Operation} presents the crossover procedure. Each individual in $pop$ is sequentially selected as Parent 1, while Parent 2 is chosen from $pop\backslash {\rm{Parent }}1$ using a 2-tournament selection strategy. After generating the offspring for $pop$ based on the crossover rate $p_c$, $offspring$ is returned.
	\begin{algorithm}[htp]
		\caption{\textbf{Function} \emph{CrossoverOperation}()}\label{PC:Crossover Operation}
		\KwIn{Problem model, $pop$.}
		\KwOut{$offspring$.}
		\LinesNumbered
		\For{$i = 1 \to popsize$}{
			${\rm{Parent }}1 \leftarrow po{p_i}$, and $offsprin{g_i} \leftarrow {\rm{Parent }}1$\;
			\If{${\rm{rand}} < p_c$}{
				${\rm{Parent 2}}$ is selected from $pop\backslash {\rm{Parent }}1$ by the 2-tournament selection\;
				\If{$n = 1$}{
					Generate $offsprin{g_i}$ by the single-point crossover;
				}
				\Else{
					Generate $offsprin{g_i}$ by the two-point crossover;
				}
			}
		}
	\end{algorithm}
	
	\subsection{\textcolor{black}{Assessment} and Ranking}\label{s36}
	
	Considering the objective function in \eqref{Eq:objective} and constraints in \eqref{Eq:constraints}, we take a COP with a minimum objective as an example. The following rules are proposed to compare two solutions ${S}$ and ${S'}$, where ${X}$ and ${Y}$ are variables of solution ${S}$, and ${X'}$ and ${Y'}$ are variables of solution ${S'}$. 
	
	\begin{itemize}
		\item  If ${S}$ is a feasible solution and ${S'}$ is infeasible, then ${S}$ is better than ${S'}$.
		
		\item As both ${S}$ and ${S'}$ are feasible solutions, we compare them based on the objective function values. If $f({X, Y}) < f({X', Y'})$, then ${S}$ is superior to ${S'}$.
		
		\item When both ${S}$ and ${S'}$ are infeasible, they are assessed by first comparing the constraints and then the objective function values. A fast non-dominated sorting strategy is employed to evaluate the constraints. If
		\begin{equation}\label{Eq:sorting strategy}
		\begin{array}{l}
		{G_i}({X}, Y) < {G_i}({X'}, Y'),{\kern 5pt}\exists i \in \{ 1, \cdots ,z\} {\kern 5pt}{\rm{and}} \\
		{G_i}({X}, Y) \le {G_i}({X'}, Y'),{\kern 5pt}\forall i \in \{ 1, \cdots ,z\},
		\end{array}
		\end{equation}
		we consider ${S}$ dominating ${S'}$, indicating ${S}$ to be better than ${S'}$. Otherwise, they have a non-dominated relationship, meaning that they are equivalent in the constraints. In this case, their objective function values are used as a secondary criterion: if $f({X, Y}) < f({X', Y'})$, then ${S}$ is better than ${S'}$.
	\end{itemize}
	
\begin{figure}[htb]
		\begin{center}\subfigure{\psfig{file=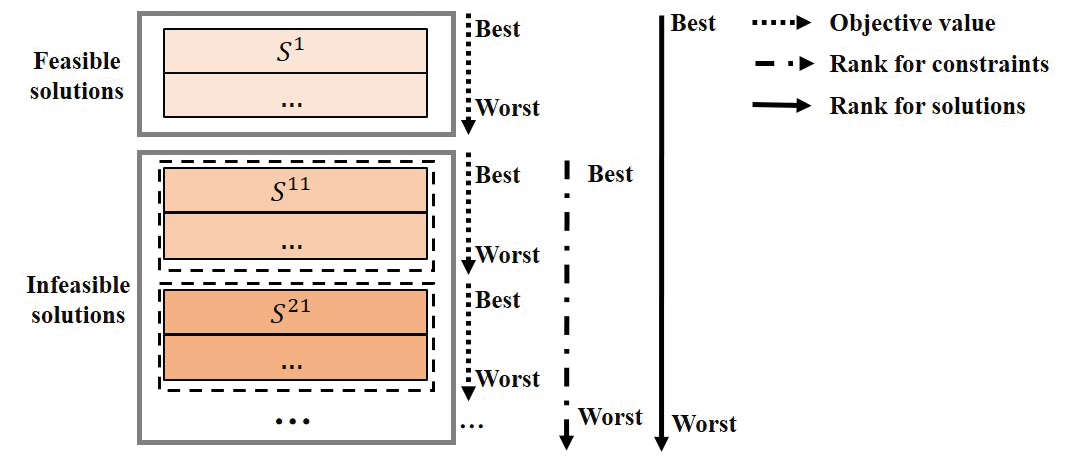, width=3.4in}}
		\end{center}
		\caption{A hierarchical ranking method for multiple solutions.}\label{Fig:ranking method}
	\end{figure}

	Based on these rules, we propose a hierarchical ranking method for multiple solutions. As shown in Fig.~\ref{Fig:ranking method}, feasible solutions are assigned better ranking values than infeasible ones. Among the feasible solutions, rankings are further determined by objective function values, with superior values leading to higher ranks. The infeasible solutions are layered based on the number of solutions they dominate, where a solution is considered superior if it dominates more other solutions. For solutions with the same level of constraint violation, further ranking is performed based on objective function values. \textcolor{black}{Due to the hierarchical ranking of solutions and dominance-based comparison for constraints, the ranking strategy can guide the search process toward high-quality feasible solutions and is suitable for COPs with diverse objectives and constraints.} 

	\subsection{Framework of Metaheuristic Algorithms}\label{s37}
	
	With the designed operators, various metaheuristic algorithms can be constructed for solving COPs. Herein, we develop two \textcolor{black}{metaheuristic} frameworks, i.e., a single-point-based framework and a population-based (GA) framework.
	
	The pseudocode for the \textcolor{black}{single-point-based} metaheuristic framework is provided in Algorithm~\ref{PC:Single-point-based Metaheuristic}. 
	\begin{algorithm}[htp]
		\caption{\textbf{Main Procedure of Single-point-based Metaheuristic Framework}}\label{PC:Single-point-based Metaheuristic}
		\KwIn{Problem model.}
		\KwOut{Best-so-far solution ${S_{{\rm{best}}}}$.}
		\LinesNumbered
		${S_{\rm{initial}}} \leftarrow $\emph{InitialSolution}()\;
		Initialize $NIter \leftarrow 0$, $Iter \leftarrow 0$, and other parameters\;
		Initialize current solution $S \leftarrow {S_{\rm{initial}}}$ and best-so-far solution ${S_{{\rm{best}}}} \leftarrow {S_{\rm{initial}}}$\;
		\While{termination condition is not satisfied}{
			Initialize the new solution set $C \leftarrow \emptyset $\;
			\While{$|C| < N$}{
				Generate a new solution ${S_{{\rm{new}}}}$\;
				$C \leftarrow C \cup {S_{{\rm{new}}}}$\;
			}
			Select the best solution ${S_{\rm{subopt}}}$ from $C$\;
			\If{${S_{\rm{subopt}}}$ is better than ${S_{{\rm{best}}}}$}{
				${S_{{\rm{best}}}} \leftarrow {S_{\rm{subopt}}}$ and $NIter \leftarrow 0$\;
			}
			\Else{
				$NIter \leftarrow NIter + 1$
			}
			$Iter \leftarrow Iter + N$\;
			Accept ${S_{{\rm{subopt}}}}$ as the current solution $S$ in terms of an acceptance criterion\;
			\If{$NIter = {\tau _s}$}{
				Perform a significant perturbation on the current solution and $NIter \leftarrow 0$\;
			}
		}
	\end{algorithm}
	
	Algorithm~\ref{PC:Single-point-based Metaheuristic} begins by generating an initial solution ${S_{\rm{initial}}}$ in line 1. The parameter $NIter$, which tracks the number of consecutive iterations without improvement, and the iteration counter $Iter$ are both initialized to 0. Subsequently, both the current solution $S$ and the best-so-far solution ${S_{{\rm{best}}}}$ are initialized to ${S_{{\rm{initial}}}}$. In each loop, a new solution set $C$ containing $N$ solutions is generated, as detailed in lines 5-8. If the best solution ${S_{{\rm{subopt}}}}$ in $C$ outperforms ${S_{{\rm{best}}}}$, the latter is updated in line 11, and $NIter$ is updated in lines 10-13. Furthermore, $S$ is updated based on a predefined acceptance criterion. The algorithm terminates and outputs ${S_{{\rm{best}}}}$ once the stopping condition is satisfied. To instantiate the single-point metaheuristic framework, we adopt VNS, TS, LNS, and SA. A detailed explanation of how these methods are implemented within the framework of Algorithm~\ref{PC:Single-point-based Metaheuristic} is provided in \textcolor{black}{the} Supplementary Material.

	Algorithm~\ref{PC:Genetic Algorithm} outlines the main procedure of GA. Lines 1-2 generate the initial population $pop$ by Algorithm~\ref{PC:Initial Solution}. Lines 6-9 focus on generating offspring for $pop$ by the crossover in Algorithms~\ref{PC:Crossover Operation} and the mutation in Algorithms~\ref{PC:Neighborhood Solution}. Since the primary goal of mutation is to enhance population diversity, a neighborhood structure is randomly selected for the mutation. Combining $pop$ and $offspring$ forms $mixpop$. In line 14, the top 10\% of individuals from $mixpop$ directly enter the next generation. The remains are selected by 2-tournament selection until the desired population size is reached. 
	
	\begin{algorithm}[htp]
		\caption{\textbf{Main Procedure of Genetic Algorithm}}\label{PC:Genetic Algorithm}
		\KwIn{Problem model.}
		\KwOut{Best-so-far solution ${S_{{\rm{best}}}}$.}
		\LinesNumbered
		\For{$i = 1 \to popsize$}{
			$po{p_i} \leftarrow $\emph{InitialSolution}()\;
		}
		Rank for $pop$ and record the best individual as $po{p_{{\rm{best}}}}$\;
		Initialize ${S_{{\rm{best}}}} \leftarrow po{p_{{\rm{best}}}}$\;
		\While{termination condition is not satisfied}{
			$offspring \leftarrow $\emph{CrossoverOperation}()\;
			\For{$i = 1 \to popsize$}{
				\If{$rand < p_m $}{
					$offsprin{g_i} \leftarrow $\emph{NeighborhoodSolution}()\;
				}
			}
			$mixpop \leftarrow pop \cup offspring$\;
			Rank for $mixpop$ and record the best individual in $mixpop$ as $mixpo{p_{{\rm{best}}}}$\;
			\If{$mixpo{p_{{\rm{best}}}}$ is better than ${S_{{\rm{best}}}}$}{
				${S_{{\rm{best}}}} \leftarrow mixpo{p_{{\rm{best}}}}$\;
			}
			Update $pop$ by a selection operator\;
		}
	\end{algorithm}
	
	\textcolor{black}{Among these methods, except for the strategies in the algorithm itself, we employ additional dynamic and adaptive mechanisms to further balance exploration and exploitation. For example, the number of removal tasks in LNS and the tabu list size in TS are dynamically adjusted over the iterations to ensure better exploration in early stages and better exploitation ability in later stages. Moreover, the Q-learning-based neighborhood selection strategy contributes to the balance by adaptively selecting promising neighborhood structures while maintaining a certain probability of random selection.}
	
	\subsection{\textcolor{black}{Complexity Analysis}}\label{s38}
	
	\textcolor{black}{The time complexities of the proposed operators are first analyzed as follows.}

	\begin{itemize}
		\item \textcolor{black}{The complexity of constructing an initial solution is ${T_1} = O(mn{n_{\max }})$, where ${n_{\max }} = \max (\{ {n_i}|i \in {\cal I}\} )$ denotes the maximum number of positions among $R$.}
		
		\item \textcolor{black}{For generating a neighborhood solution, the complexity is ${T_2} = \max ({n_{\max }},m{n_{\max }},1,m{n_{\max }},m{n_{\max }},2{n_{\max }}) = O(m{n_{\max }})$.}
		
		\item \textcolor{black}{For destroying and repairing a solution, the complexity is ${T_3} = nd \cdot {n_{\max }} + mn{n_{\max }} = O(mn{n_{\max }})$.}
		
		\item \textcolor{black}{When executing the crossover operator, the complexity is ${T_4} = mn + mn{n_{\max }} = O(mn{n_{\max }})$.}
		
		\item \textcolor{black}{When ranking for ${N_S}$ solutions, the complexity is ${T_5} = O(zN_S^2)$, where $z$ is the number of constraints.}
		
	\end{itemize}

	\textcolor{black}{As shown above, the complexity of the ranking method depends on the number of constraints and solutions to be ranked. Moreover, the complexity for others grows linearly with the number of resources $m$, the number of tasks $n$, and the maximum number of positions $n_{\max}$. }

	\textcolor{black}{For the specific metaheuristic algorithms, the complexity of the single-point-based methods with the neighborhood structures, like SA, TS, and VNS, is}
	\begin{equation}\label{Eq:Complexity 1}
	\begin{array}{l}
	\color{black}{{T_6} = {T_1} + {\rm{MaxIter}} \cdot {\rm{(}}N \cdot {T_2}{\rm{ + }}{T_5}{\rm{) }}} \\\color{black}{ = O({\rm{MaxIter}} \cdot N \cdot m{n_{\max }}).}
	\end{array}
	\end{equation}
	
	\textcolor{black}{For the single-point-based methods with the destroy-and-repair operator, i.e., LNS, the complexity is}
	\begin{equation}\label{Eq:Complexity 2}
	\begin{array}{l}
	\color{black}{{T_7} = {T_1} + {\rm{MaxIter}} \cdot {\rm{(}}N \cdot {T_3}{\rm{ + }}{T_5}{\rm{) }}} \\ \color{black}{O({\rm{MaxIter}} \cdot N \cdot mn{n_{\max }}).}
	\end{array}
	\end{equation}
	
	\textcolor{black}{Concerning GA shown in Algorithm~\ref{PC:Genetic Algorithm}, the complexity is}
	\begin{equation}\label{Eq:Complexity 3}
	\begin{array}{l}
	\color{black}{{T_8} = popsize \cdot {T_1} + {\rm{MaxIter}} \cdot {\rm{(}}popsize \cdot ({T_4}{\rm{ + }}{T_2}{\rm{) + }}{T_5}{\rm{)}}}\\
	\color{black}{{\rm{ = }}O( {\rm{MaxIter}} \cdot popsize \cdot mn{n_{\max }}).}
	\end{array}
	\end{equation}
	
	\textcolor{black}{Except for problem size, the computational complexity of a given algorithm also grows linearly with the number of iterations $\rm{MaxIter}$, neighborhood size $N$, and population size $popsize$. Therefore, such algorithms can efficiently solve COPs with appropriate problem scale and algorithm parameter settings. Moreover, given the same number of function evaluations, the computational complexities of GA and LNS are comparable, while the single-point methods based on neighborhood structures have lower complexity.}

	\section{Experimental Study}\label{s4}
	
	\subsection{Test Suit and Experimental Setting}\label{s41}
	
	To evaluate the effectiveness of our proposal, 10 COPs are selected, including multi-depot vehicle routing problem with time windows (MDVRPTW) \cite{zhen2009hybrid}, vehicle routing problem with pickup and delivery (VRPPD) \cite{van2001comparing}, generalized assignment problem (GAP) \cite{osman1995heuristics}, capacitated facility location problem (CFLP) \cite{sridharan1995capacitated}, bin packing problem with conflicts (BPPC) \cite{muritiba2010algorithms}, generalized bin packing problem (GBPPI) \cite{baldi2019generalized}, quadratic multiple knapsack problem (QMKP) \cite{galli2021polynomial}, high school timetabling problem (HTS) \cite{saviniec2017effective}, job shop scheduling problem (JSSP) \cite{taillard1993benchmarks}, and graph coloring problem (GC) \cite{fleurent1996genetic}. This selection encompasses classical problems like JSSP and GC, generalized problems such as GAP, CFLP, and GBPPI, and variants of classical problems like MDVRPTW, VRPPD, BPPC, QMKP, and HTS. The introduction and resource-centered modeling for each type of COP are provided in the Supplementary Material due to page limitations.
	
	\textcolor{black}{The objective and gap values obtained by each algorithm on each instance are used as indicators to evaluate algorithm performance. To calculate the gap values, the lower bound for each problem is obtained using GUROBI with default settings within 3,600 seconds. Moreover, the number of evaluations is either unavailable or not directly controllable for OR-TOOLS, SCIP, and GUROBI. In practice, users are typically more concerned with the quality of solutions obtained within a fixed time limit than with the number of evaluations. Therefore, to ensure a fair and meaningful comparison, all algorithms are allocated the same maximum runtime for each instance, as specified in the Supplementary Material. In addition, each algorithm is run 5 times on each problem for reliability.}
	
	\subsection{\textcolor{black}{Parameter Analysis and Setting}}\label{s42}
	
	\textcolor{black}{Apart from dynamic parameters such as the number of removed tasks in LNS and the size of the tabu list in TS, Table S-10 presents other parameters and their possible values for the 5 algorithms in REMS. We employ the Taguchi method to study the impact of these parameters on algorithm performance. 60 instances from different problem types are randomly selected for the experiment. Each parameter combination is independently tested 5 times on each instance. Table S-11 presents the signal-to-noise (S/N) ratios for each parameter, calculated based on the mean gap values for the corresponding parameter combinations.}
	
	\textcolor{black}{According to Table S-11, although the parameters introduced by the Q-learning-based selection strategy possess different impacts on different algorithms, they generally have a low impact, indicating the robustness of the selection mechanism. Other specific algorithm parameters, like the cooling rate of SA as well as crossover and mutation rates of GA, have a larger impact on the corresponding algorithms. This observation motivates future work on developing more effective algorithms for REMS by parameter tuning. Based on Table S-11, the final parameter settings adopted in this study are presented in Table S-12.}

	\subsection{Compare Metaheuristic Algorithms in REMS on Different Problems}\label{s43}
	
	The subsection compares 5 metaheuristic algorithms in REMS across 10 problem types, evaluating their effectiveness in solving various COPs and identifying which algorithm performs best for each instance and problem type. \textcolor{black}{The detailed experimental results are presented in Tables S-11 to S-20 of Supplementary Material, including the best and mean objective function values and gap values over 5 runs. }
	
	\subsubsection{\textcolor{black}{Performance Comparison on Specific Instances}}\label{s431}
	
	In addition to the results presented in Tables S-11 to S-20, \textcolor{black}{Fig. S-1} illustrates the radar chart for the mean gap values obtained by 5 metaheuristic algorithms in REMS across 10 problems. In each radar chart, a point closer to the center of the circle indicates better mean gap values achieved by an algorithm for a given instance.
	
	The results presented in Fig. S-1 and Tables S-11 to S-20 reveal that the performance of a metaheuristic algorithm in REMS varies both across different problem types and among instances within the same type. The observation is consistent with the No Free Lunch (NFL) theorem, which states that no single algorithm can outperform all others across all problems.

	\subsubsection{\textcolor{black}{Overall Performance Analysis of Algorithms}}\label{s432}
	
\begin{figure}[htb]
		\begin{center}
			\subfigure{\psfig{file=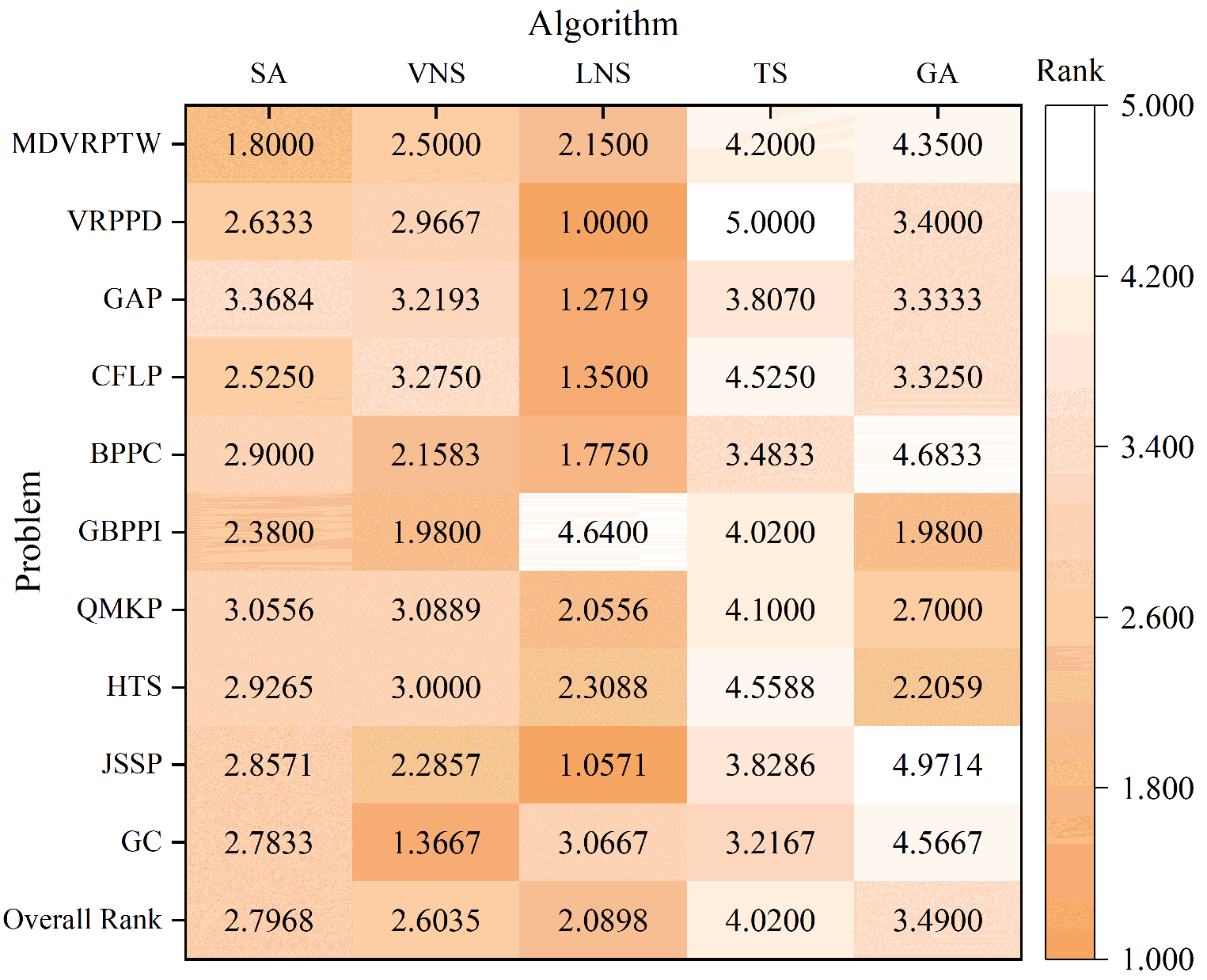, width=3.4in}}\hspace{-0.2in}
		\end{center}\vspace{-5mm}
		\caption{\textcolor{black}{Heatmap diagram for 5 algorithms in REMS on 10 problems.}}\label{Fig:Heatmap diagram for 5 metaheuristic algorithms}
	\end{figure}

	To evaluate the overall performance, Fig.~\ref{Fig:Heatmap diagram for 5 metaheuristic algorithms} provides the heatmap diagram of the Friedman test results conducted on the mean gap values across different problem types. In the heatmap, darker colors correspond to better rankings.
	
	Based on Fig.~\ref{Fig:Heatmap diagram for 5 metaheuristic algorithms}, despite the different ranking values across problem types, LNS achieves the best rankings for \textcolor{black}{6} problem types and obtains the best overall ranking. In contrast, TS exhibits the worst rankings in \textcolor{black}{5} problem types based on the mean gap indicators, which has the poorest overall ranking.
	
	To compare the differences among algorithms, \textcolor{black}{Fig. S-2} presents the critical difference (CD) \textcolor{black}{plots} of mean gap values for 5 metaheuristic algorithms on 10 problems, with a significance level of 0.05. The plot reveals whether pairs of algorithms are statistically significant. Specifically, if two algorithms are connected by a horizontal line, their performance differences are not statistically significant. In \textcolor{black}{terms} of the results, Table~\ref{tab:1} further summarizes the pairwise comparison results, indicating the number of problems where one algorithm is significantly better ($+$), significantly worse ($ - $), or shows no significant difference ($ \approx $) compared to another.
	
	\aboverulesep=0pt \belowrulesep=0pt
	\begin{table}[htbp]
		\centering
		\caption{Number of problems where an algorithm is significantly better, significantly worse, or shows no significant difference compared to another algorithm ($ + \backslash  - \backslash  \approx $).}
		\setlength{\tabcolsep}{2.5mm}{
			\color{black}{\begin{tabular}{cccccc}
				\toprule
				Algorithm & SA    & VNS   & LNS   & TS    & GA \\
				\midrule
				SA \textit{VS} & /     & 0\textbackslash{}1\textbackslash{}9 & 1\textbackslash{}6\textbackslash{}3 & 6\textbackslash{}0\textbackslash{}4 & 4\textbackslash{}0\textbackslash{}6 \\
				VNS \textit{VS} & 1\textbackslash{}0\textbackslash{}9 & /     & 2\textbackslash{}5\textbackslash{}3 & 9\textbackslash{}0\textbackslash{}1 & 4\textbackslash{}0\textbackslash{}6 \\
				LNS \textit{VS} & 6\textbackslash{}1\textbackslash{}3 & 5\textbackslash{}2\textbackslash{}3 & /     & 8\textbackslash{}0\textbackslash{}2 & 7\textbackslash{}1\textbackslash{}2 \\
				TS \textit{VS} & 0\textbackslash{}6\textbackslash{}4 & 0\textbackslash{}9\textbackslash{}1 & 0\textbackslash{}8\textbackslash{}2 & /     & 3\textbackslash{}5\textbackslash{}2 \\
				GA \textit{VS} & 0\textbackslash{}4\textbackslash{}6 & 0\textbackslash{}4\textbackslash{}6 & 1\textbackslash{}7\textbackslash{}2 & 5\textbackslash{}3\textbackslash{}2 & / \\
				\bottomrule
			\end{tabular}%
		}
		}
		\label{tab:1}%
	\end{table}%
	
	
	From the perspective of each problem, the differences vary from problem to problem as well. \textcolor{black}{For instance, in MDVRPTW, no significant differences are observed among SA, LNS, and VNS, or between TS and GA. Regarding GBPPI, GA, VNS, and SA exhibit similar performance, as do LNS and TS. Regarding JSSP, LNS significantly outperforms other algorithms and no significant differences are found between VNS and SA or between SA and TS. For the GC problem, the performance differences among SA, TS, and LNS are not statistically significant.}
	
	According to Table~\ref{tab:1}, LNS consistently outperforms or shows comparable performance to the other algorithms across the majority of problem types. Moreover, the results obtained by TS are either significantly worse than or not significantly different from those of the other algorithms. Furthermore, pairwise comparisons between the algorithm pairs, \textcolor{black}{VNS and SA, GA and VNS, and SA and GA,} reveal no significant difference in more than half of the problem types.

	\subsubsection{\textcolor{black}{Convergence Analysis}}\label{s433}
	
	To explore the convergence behavior, we provide the convergence curves of 5 metaheuristic algorithms on 4 typical instances over a typical run for the 10 problem types, shown in Fig. S-3 to S-12. A typical run refers to one whose objective value is closest to the mean objective value obtained over 5 runs.
	
	As shown in Figs. S-3 to S-12, all 5 algorithms in REMS can effectively converge. Among them, GA exhibits the slowest convergence speed. The 4 single-point-based algorithms converge more quickly, with LNS consistently identifying better solutions during the iterations for most problems. 
	
	\subsubsection{\textcolor{black}{Summary}}\label{s434}
	
	From the above experimental study, REMS can effectively model and solve 10 problem types, and the performance of constructed algorithms within REMS varies across different problem types, and even among different instances of the same problem type. Overall, LNS gains the best results and TS exhibits the worst results among the 5 algorithms. Regarding convergence, the population-based algorithm, i.e., GA, exhibits slower convergence speeds than the other four single-point-based algorithms.
	

	\subsection{Compare Metaheuristic Algorithms with Other Solvers}\label{s44}
	
	This subsection compares 5 metaheuristic algorithms in REMS with several promising solvers, including CP-SAT and Routing solvers from OR-TOOLS, as well as SCIP and GUROBI. Notably, the Routing solver, specifically designed for vehicle routing problems, is applied only to MDVRPTW and VRPPD herein. Furthermore, CP-SAT is not suitable for problems with continuous variables and is therefore excluded from solving MDVRPTW and VRPPD. \textcolor{black}{Detailed experimental results are presented in Tables S-11 to S-20. }
	
	\subsubsection{\textcolor{black}{Performance Comparison on Specific Instances}}\label{s441}
	
	Besides the results in Tables S-11 to S-20, the radar chart of the mean gap values obtained by 8 algorithms on 10 problems is presented in Fig. \textcolor{black}{S-13}. 
	
	According to Tables S-11 to S-20 and Fig. \textcolor{black}{S-13}, the results reveal that compared to the exact algorithm-based solvers, i.e., GUROBI and SCIP, 5 metaheuristic algorithms in REMS obtain better solutions, especially for large-scale instances and complex problems. Notably, these algorithms outperform GUROBI and SCIP on all instances of MDVRPTW and VRPPD. \textcolor{black}{When it comes to BPPC, GBPPI, and GC, GUROBI fails to find any feasible solutions for several larger-scale instances. Meanwhile, SCIP can obtain feasible solutions for most BPPC and GC instances, but these are not superior to those found by the metaheuristics in REMS. Regarding QMKP, the emergence of the quadratic objective function makes the metaheuristic algorithms in REMS outperform GUROBI and SCIP in most instances.} Concerning HTS, which involves complex objectives and constraints, the 5 metaheuristic algorithms in REMS consistently yield better solutions than GUROBI and SCIP for the majority of instances. Furthermore, the solutions provided by GUROBI and SCIP are relatively stable over different runs, but some randomness remains due to some heuristic rules being employed in them.
	
	In comparison with OR-TOOLS, the experimental results obtained by the algorithms in REMS are worse in most instances. On the one hand, OR-TOOLS exhibits good performance on some relatively large-scale instances, including GC21-30, BPPC11-20, and 51-60. This can be attributed to the effective incorporation of various heuristic methods and promising strategies. On the other hand, it also performs well in problems that benefit from its specific modeling approach and algorithmic design, such as VRPPD and JSSP. However, the 5 algorithms in REMS are mainly based on classical metaheuristic algorithms. Therefore, the performance of the algorithms in REMS is not as good as OR-TOOLS in most instances. This phenomenon inspires us to improve the performance of the algorithms in REMS by integrating advanced methods in future work.
	
	Nevertheless, an interesting observation is that the metaheuristic algorithms embedded in REMS have the potential to locate better solutions for several problems with complex objectives and constraints, including MDVRPTW, HTS, and QMKP. Regarding MDVRPTW, OR-TOOLS fails to locate feasible solutions except for MDVRPTW07, 09, 16, 17, and 19, whereas 5 metaheuristic algorithms in REMS can realize it. Additionally, due to the complexity of HTS and QMKP, the solutions located by OR-TOOLS are inferior to those obtained by REMS in most instances.

	\subsubsection{\textcolor{black}{Overall Performance Analysis of Algorithms}}\label{s442}
	
\begin{figure}[htb]
		\begin{center}
			\subfigure{\psfig{file=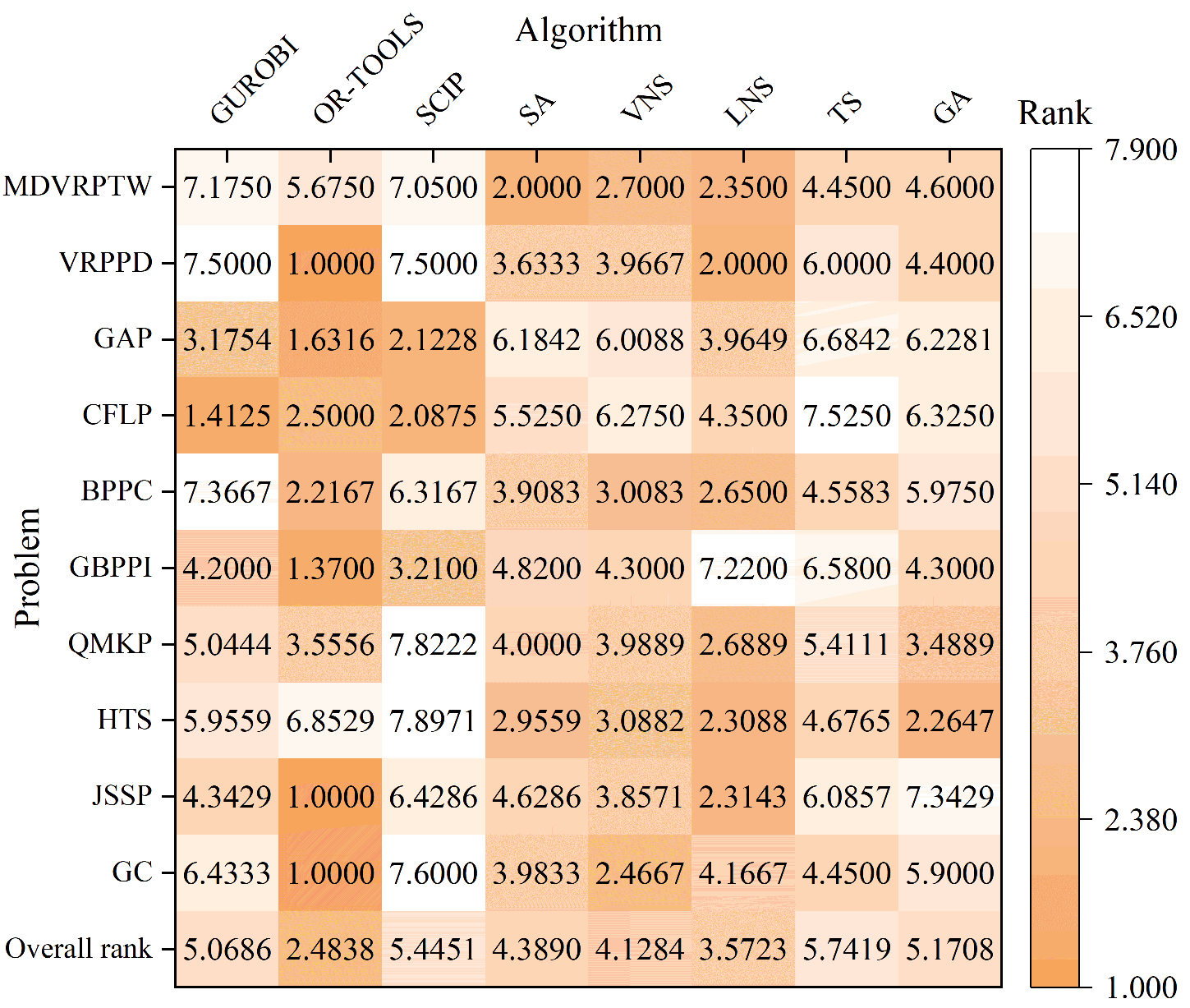, width=3.4in}}\hspace{-0.2in}
		\end{center}\vspace{-5mm}
		\caption{ Heatmap diagram for 8 algorithms on 10 problems}\label{Fig:Heatmap diagram for 8 algorithms}
	\end{figure}

	To compare the overall performance of the 8 algorithms, \textcolor{black}{Fig.~\ref{Fig:Heatmap diagram for 8 algorithms} presents the heatmap diagram of Friedman ranking values based on the mean gap across 10 problems.}
	
	Compared to GUROBI and SCIP, our proposal has the capacity to acquire better ranking values on most problem types. In particular, \textcolor{black}{5 metaheuristic algorithms embedded in REMS can provide better mean gap ranking values on MDVRPTW, VRPPD, BPPC, HTS, and GC. When solving QMKP, SA, VNS, LNS, and GA in REMS also achieve superior rankings compared to GUROBI, and all 5 metaheuristic algorithms of REMS outperform SCIP.} 
	
	OR-TOOLS can obtain better ranking values on 7 problems, excluding MDVRPTW, QMKP, and HTS. For MDVRPTW and HTS, all metaheuristic algorithms in REMS gain a better ranking value than OR-TOOLS. For QMKP, LNS and GA outperform OR-TOOLS.
	
	To further compare the difference among algorithms, Fig. \textcolor{black}{S-14} presents the CD \textcolor{black}{plots} of mean gap for 8 algorithms on 10 problems. Table~\ref{tab:2} summarizes the number of problems where GUROBI, OR-TOOLS, or SCIP is significantly better, significantly worse, or no significant difference compared to the metaheuristic algorithms of REMS. 
	
	\aboverulesep=0pt \belowrulesep=0pt
	\begin{table}[htbp]
		\centering
		\caption{Number of problems where GUROBI, OR-TOOLS, or SCIP is significantly better, significantly worse, or shows no significant difference compared to another metaheuristic algorithm in REMS ($ +\backslash-\backslash\approx $).}
		\setlength{\tabcolsep}{2.5mm}{
			\color{black}{\begin{tabular}{cccccc}
				\toprule
				Algorithm & SA    & VNS   & LNS   & TS    & GA \\
				\midrule
				GUROBI \textit{VS} & 2\textbackslash{}5\textbackslash{}3 & 2\textbackslash{}5\textbackslash{}3 & 2\textbackslash{}7\textbackslash{}1 & 3\textbackslash{}3\textbackslash{}4 & 3\textbackslash{}4\textbackslash{}3 \\
				OR-TOOLS \textit{VS} & 6\textbackslash{}3\textbackslash{}1 & 5\textbackslash{}3\textbackslash{}2 & 4\textbackslash{}2\textbackslash{}4 & 8\textbackslash{}1\textbackslash{}1 & 7\textbackslash{}1\textbackslash{}2 \\
				SCIP \textit{VS} & 3\textbackslash{}7\textbackslash{}0 & 2\textbackslash{}7\textbackslash{}1 & 3\textbackslash{}7\textbackslash{}0 & 3\textbackslash{}5\textbackslash{}2 & 2\textbackslash{}5\textbackslash{}3 \\
				\bottomrule
			\end{tabular}%
		}
		}
		\label{tab:2}%
	\end{table}%
	
	Based on Fig.~\textcolor{black}{S-14} and Table~\ref{tab:2}, the metaheuristic algorithms in REMS are either significantly better than or exhibit no significant difference compared to GUROBI and SCIP on more than 50\% of problem types. Specifically, SA, VNS, LNS, TS, and GA are significantly better than or equal to GUROBI on \textcolor{black}{8, 8, 8, 7, and 7} problem types, respectively. Furthermore, SA, VNS, LNS, TS, and GA achieve comparable or superior performance to SCIP on \textcolor{black}{7, 8, 7, 7, and 8} problem types, respectively.
	
	Additionally, LNS consistently outperforms or shows no significant difference when compared to OR-TOOLS across most problems. Especially, OR-TOOLS is significantly worse than GA, VNS, LNS, and SA for MDVRPTW. Moreover, LNS shows no significant difference with OR-TOOLS when solving VRPPD, BPPC, QMKP, and JSSP.


	\subsubsection{\textcolor{black}{Convergence Analysis}}\label{s443}
	
	To compare the convergence behavior of metaheuristic algorithms embedded in REMS with other algorithms, Figs. S-15 to S-24 illustrate the convergence curve for 8 algorithms on 4 typical instances during a typical run for each problem type. 
	
	According to these figures, in some instances, especially for large-scale or complex instances, GUROBI, OR-TOOLS, and SCIP may struggle to effectively converge or converge to worse solutions than the metaheuristic methods in REMS. For instance, when solving MDVRPTW09, OR-TOOLS requires a longer time to find a good solution, SCIP converges \textcolor{black}{to} a worse solution, and GUROBI fails to locate any feasible solutions.
	
	
	\subsubsection{\textcolor{black}{Summary}}\label{s444}
	
	Overall, compared with exact algorithm-based solvers, i.e., GUROBI and SCIP, REMS outperforms these solvers on large-scale instances and complex COPs, including MDVRPTW, VRPPD, BPPC, QMKP, HTS, JSSP, and GC. Additionally, REMS exhibits superior performance to OR-TOOLS, when solving certain complex COPs, such as MDVRPTW, QMKP, and HTS.
	
	\subsection{\textcolor{black}{Effectiveness of Q-Learning-Based Selection Strategy}}\label{s45}
	
	\textcolor{black}{To assess the effectiveness of the Q-learning-based neighborhood selection method in Section \ref{s33}, the following comparison methods are employed: (a) qVNS: VNS with the Q-learning-based selection; (b) aVNS: VNS with the adaptive selection based on historical success rates; and (c) rVRS: VNS with random selection. Detailed results about the mean objective and gap values are presented in Tables S-23 to S-32.}
	
	\subsubsection{\textcolor{black}{Performance Comparison on Specific Instances}}\label{s451}
	
	\textcolor{black}{In terms of Tables S-23 to S-32, qVNS demonstrates superior performance over aVNS and rVNS in most instances. Specifically, qVNS yields better results on 87.79\% and 97.96\% of the instances, compared with aVNS and rVNS, respectively.}

	\subsubsection{\textcolor{black}{Overall Performance Analysis of Algorithms}}\label{s452}
	
	\textcolor{black}{Tables~\ref{Frideman test} and~\ref{tab:Wilcoxon signed-rank 0} summarize the Friedman test and Wilcoxon signed-rank test results based on the mean gap values for the three compared methods on 10 problem types, respectively.}
	
	\aboverulesep=0pt \belowrulesep=0pt
	\begin{table}[htbp]
		\centering
		\caption{\textcolor{black}{Friedman test for qVNS, aVNS, and rVNS.}}
		\setlength{\tabcolsep}{0.1mm}{
			\color{black}{\begin{tabular}{ccccccccccc}
				\toprule
				Method & \tabincell{c}{MDVRP\\TW} & \tabincell{c}{VRPPD} & GAP & CFLP & BPPC & \tabincell{c}{GBP\\PI} & QMKP & HTS & JSSP & GC \\
				\midrule
				qVNS  & \textbf{1.100 } & \textbf{1.133 } & \textbf{1.123 } & \textbf{1.025 } & \textbf{1.217 } & \textbf{1.120 } & \textbf{1.267 } & \textbf{1.147 } & \textbf{1.171 } & \textbf{1.133 } \\
				aVNS  & 1.950  & 1.900  & 1.974  & 1.975  & 2.017  & 1.880  & 1.933  & 1.882  & 1.829  & 1.933  \\
				rVNS  & 2.950  & 2.967  & 2.904  & 3.000  & 2.767  & 3.000  & 2.800  & 2.971  & 3.000  & 2.933  \\
				\bottomrule
			\end{tabular}%
		}
		}
		\label{Frideman test}%
	\end{table}%

	\aboverulesep=0pt \belowrulesep=0pt
	\begin{table}[htbp]
		\centering
		\caption{\textcolor{black}{Wilcoxon signed-rank results for comparing qVNS with others at a 0.05 significance level, where the $p$ values that are less than 0.05 are bolded.}}
		\setlength{\tabcolsep}{1.8mm}{
			\color{black}{\begin{tabular}{ccccccc}
				\toprule
				\multicolumn{1}{c}{\multirow{2}{*}{Problem}} & \multicolumn{3}{c}{qVNS \textit{VS} aVNS} & \multicolumn{3}{c}{qVNS \textit{VS} rVNS} \\
				\cmidrule{2-7}    \multicolumn{1}{c}{} &  ${R^ + }$    &  ${R^ - }$    & $p$-value & ${R^ + }$    &  ${R^ - }$    & $p$-value \\
				\midrule
				MDVRPTW & 200.0  & 10.0  & \textbf{3.63E-04} & 210.0  & 0.0   & \textbf{8.20E-05} \\
				VRPPD & 455.0  & 10.0  & \textbf{5.00E-06} & 465.0  & 0.0   & \textbf{2.00E-06} \\
				GAP   & 1574.5  & 21.5  & \textbf{3.55E-10} & 1590.5  & 5.5   & \textbf{1.82E-10} \\
				CFLP  & 819.0  & 1.0   & \textbf{3.64E-12} & 820.0  & 0.0   & \textbf{1.82E-12} \\
				BPPC  & 1744.0  & 86.0  & \textbf{8.74E-09} & 1812.5  & 17.5  & \textbf{1.71E-10} \\
				GBPPI & 1141.0  & 134.0  & \textbf{1.00E-06} & 1275.0  & 0.0   & \textbf{1.78E-15} \\
				QMKP  & 864.5  & 170.5  & \textbf{8.60E-05} & 1035.0  & 0.0   & \textbf{5.68E-14} \\
				HTS   & 466.0  & 129.0  & \textbf{3.86E-03} & 595.0  & 0.0   & \textbf{1.16E-10} \\
				JSSP  & 545.0  & 85.0  & \textbf{1.55E-04} & 630.0  & 0.0   & \textbf{5.82E-11} \\
				GC    & 391.5  & 73.5  & \textbf{9.75E-04} & 465.0  & 0.0   & \textbf{2.00E-06} \\
				\bottomrule
			\end{tabular}%
		}
		}
		\label{tab:Wilcoxon signed-rank 0}%
	\end{table}%
	
	\textcolor{black}{As evidenced by the better ranking values of qVNS in Table~\ref{Frideman test}, the Q-learning-based method enables VNS to achieve superior overall results in each problem type. Furthermore, the fact that ${R^+}$ exceeds ${R^-}$ and the $p$ values are below 0.05 indicates that qVNS significantly outperforms the others across all problem types. In addition, regarding each problem type, the Q-learning-based selection method enhances the performance of VNS to different extents.}

	\subsubsection{\textcolor{black}{Convergence Analysis}}\label{s453}
	
	\textcolor{black}{Figs. S-25 to S-34 illustrate the convergence curves of objective values on 4 instances during a typical run for each problem type. The curves indicate that qVNS consistently converges to better solutions. Notably, although qVNS may perform similarly to or worse than aVNS in the early stages, it gradually learns effective selection strategies, leading to a better performance in the later stages. In addition, aVNS generally can converge to a better solution than rVNS, where the convergence process of the latter is unstable and tends to be trapped in local optima.}

	\subsubsection{\textcolor{black}{Summary}}\label{s454}
	
	\textcolor{black}{When integrating the Q-learning method to select neighborhood structures, an algorithm enables superior solutions and better convergence performance than the random selection and the historical search information-based selection for solving various COPs.}

	\section{Conclusion}\label{s5}
	
	\textcolor{black}{By leveraging more general problem domain knowledge, we propose a resource-centered modeling and solving framework (REMS) to formulate diverse COPs into a unified paradigm and improve the generality and adaptability of metaheuristics.} To evaluate the effectiveness of REMS, an extensive experimental study is conducted across 10 COPs, including routing, location, loading, assignment, scheduling, and graph coloring problems. The results manifest that REMS has the capacity to express and effectively solve diverse COPs without any specific design. Moreover, REMS \textcolor{black}{exhibits} superior performance \textcolor{black}{to} GUROBI and SCIP for large-scale instances and complex problems. REMS can also provide better performance compared with OR-TOOLS in solving several complex problems. 
	
	However, the COPs addressed in this paper do not consider multiple conflicting objectives, dynamic environments, or multiple sub-problems. Addressing these problems is a critical focus in future research. Additionally, several metaheuristic methods with better generality and adaptability for general COPs are embedded in REMS herein. However, their efficiency may be hindered by the problem nature, parameter settings, algorithms themselves, and other factors. \textcolor{black}{Benefiting from the generalization of neural networks and the self-learning capability of RL, a promising direction lies in applying them to improve the effectiveness of REMS. For example, they can select suitable operators and algorithm parameters or construct efficient operators by further exploring specific problem characteristics. Moreover, Large Language Models (LLMs) can be applied to support better problem description and more effective algorithm design \cite{romera2024mathematical, liu2024evolution, yang2025heuragenix} by exploring the problem and algorithm knowledge from the resource-task perspective. Furthermore, both ensemble strategies \cite{wu2019ensemble}, which combine the advantages of different methods, and problem reduction methods \cite{song2024exact, sun2019using}, designed to simplify problem complexity, are also potential improvements.}

	
	%

	



	\ifCLASSOPTIONcaptionsoff
	\newpage
	\fi

	\bibliographystyle{IEEEtran}
	\bibliography{IEEEabrv,mybib}

	\vspace{-12mm}
	
	\begin{IEEEbiography}[{\includegraphics[width=0.8in,height=1.1in, clip, keepaspectratio]{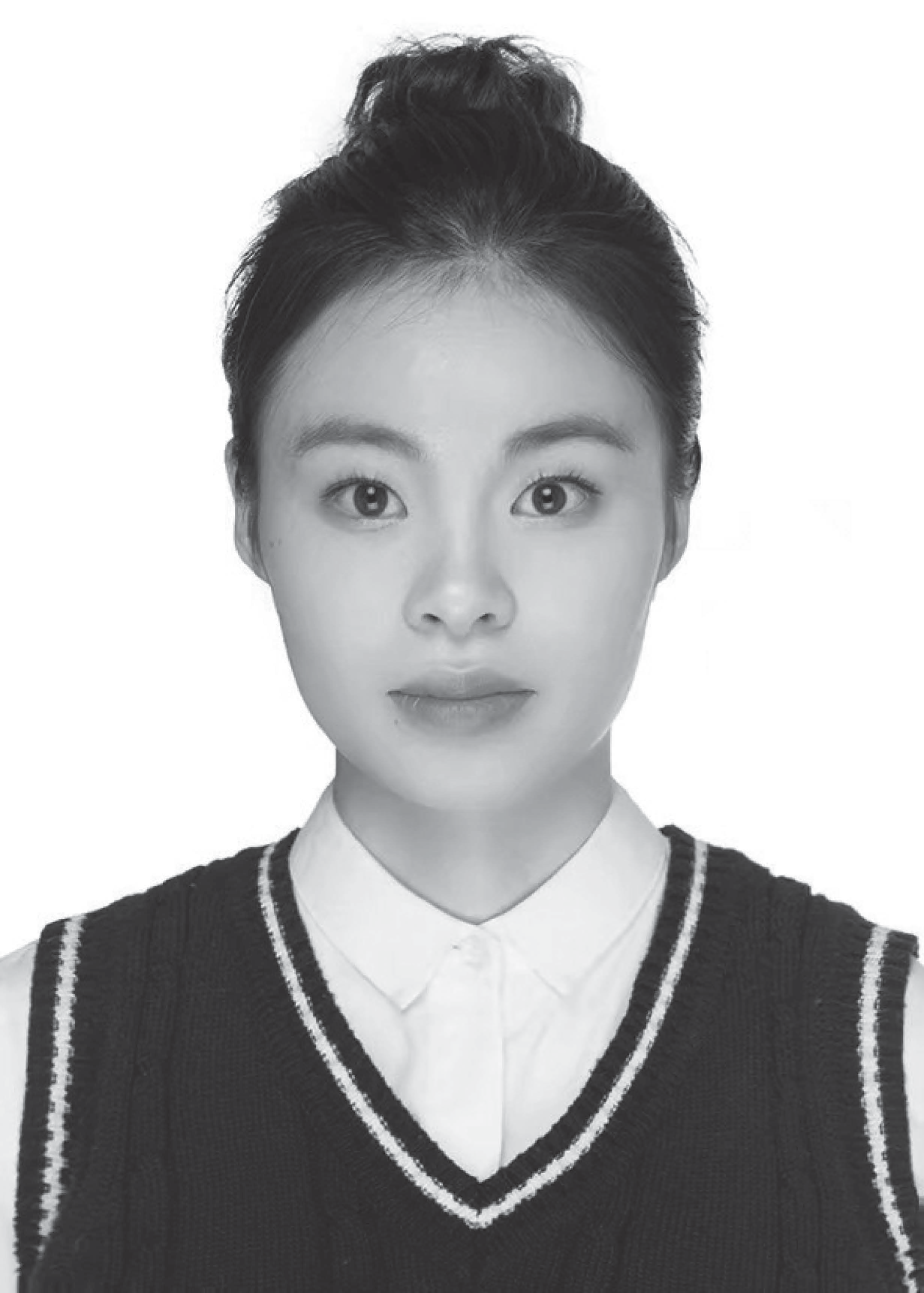}}] {Aijuan Song} received her B.E. and M.E. degrees from Central South University, China, in 2019 and 2022. Currently, she is working toward her Ph.D. degree at the School of Traffic \& Transportation Engineering, Central South University. Her research interests include Computational Intelligence, Planning and Scheduling.
	\end{IEEEbiography}
	
	\vspace{-12mm}
	
	\begin{IEEEbiography}[{\includegraphics[width=1.0in,height=1.2in, clip, keepaspectratio]{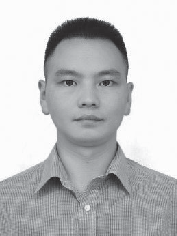}}] {Guohua Wu} received the B.S. degree in Information Systems and Ph.D degree in Operations Research from National University of Defense Technology, China, in 2008 and 2014, respectively. During 2012 and 2014, he was a visiting Ph.D student at University of Alberta, Edmonton, Canada. He is currently a Professor at the School of Automation, Central South University, Changsha, China. His current research interests include Planning and Scheduling, Computational Intelligence and Machine Learning. He serves as an Associate Editor of \emph{Information Sciences}, and an Associate Editor of \emph{Swarm and Evolutionary Computation Journal}.  
	\end{IEEEbiography}
	
	

\end{document}